\documentclass[letterpaper]{article} % DO NOT CHANGE THIS
\usepackage{aaai24}  % DO NOT CHANGE THIS
\usepackage{times}  % DO NOT CHANGE THIS
\usepackage{helvet}  % DO NOT CHANGE THIS
\usepackage{courier}  % DO NOT CHANGE THIS
\usepackage[hyphens]{url}  % DO NOT CHANGE THIS 
\usepackage{graphicx} % DO NOT CHANGE THIS
\usepackage{algpseudocode}
\usepackage{amsmath}
\newcommand{\algcomment}[1]{\textcolor{green!70!black}{\texttt{\# #1}}}

\usepackage{natbib}  % DO NOT CHANGE THIS AND DO NOT ADD ANY OPTIONS TO IT
\usepackage{caption} % DO NOT CHANGE THIS AND DO NOT ADD ANY OPTIONS TO IT
\usepackage{algorithm}
\usepackage{booktabs}
\usepackage{multirow}
\usepackage{amsfonts}
\usepackage{mathrsfs}  % \mathscr
\usepackage[capitalize,noabbrev]{cleveref} % \cref, ,noabbrev
\usepackage{tabularx}
\usepackage{adjustbox}
\usepackage{makecell}
\usepackage{subfig} % Commented out as it's often redundant with subcaption
\usepackage{threeparttable}
\usepackage{array}
\usepackage{xspace}
\usepackage{verbatim} 
\usepackage{nicefrac}
\usepackage{comment}
\usepackage[table,xcdraw]{xcolor} 
\usepackage{dsfont}
\usepackage{xr}
\usepackage{subfiles}
\usepackage{enumitem}
\usepackage{mdframed}
\usepackage{tcolorbox}
\usepackage[T1]{fontenc}
\usepackage{float}

\DeclareCaptionFormat{myformat}{Code #2 #3}
\definecolor{sky}{cmyk}{0.65,.0,.0,0}
\usepackage{listings}
\newtheorem{definition}{Definition}

\DeclareMathOperator{\diam}{diam}

\newcommand{\stdv}[1]{\scriptsize$\pm$#1}

\usepackage{newfloat}
\usepackage{listings}
\DeclareCaptionStyle{ruled}{labelfont=normalfont,labelsep=colon,strut=off} % DO NOT CHANGE THIS
\floatstyle{ruled}
\newfloat{listing}{tb}{lst}{}
\floatname{listing}{Listing}
\newcommand{\suppletitle}[1]{%
\vbox{
    \centering
    {\LARGE\bf #1\par}
    \vskip 0.1in
  }
}

\title{Unveiling the Significance of Toddler-Inspired Reward Transition \\in Goal-Oriented Reinforcement Learning}

\author{
    Junseok Park\textsuperscript{\rm 1},
    Yoonsung Kim\textsuperscript{\rm 1}\footnote{Equal Contributions.},
    Hee Bin Yoo\textsuperscript{\rm 1}\textsuperscript{*},
    Min Whoo Lee\textsuperscript{\rm 1},
    Kibeom Kim\textsuperscript{\rm 1},
    Won-Seok Choi\textsuperscript{\rm 1},
    Minsu Lee\textsuperscript{\rm 1,2}\footnote{Corresponding Authors.},
    Byoung-Tak Zhang\textsuperscript{\rm 1,2,}\textsuperscript{†}
}
\affiliations{
    \textsuperscript{\rm 1}Seoul National University\\
    \textsuperscript{\rm 2}AI Institute of Seoul National University (AIIS)\\
    \{jspark, yskim, hbyoo, mwlee, kbkim, wchoi, mslee, btzhang\}@bi.snu.ac.kr
}

\usepackage{bibentry}
\begin{document}

\maketitle

\begin{abstract}

%-------------UPGRADE 버전 BY 준석
%We highlight notable parallels between the learning behaviors observed in toddlers and the strategies employed in reward shaping for goal-oriented Reinforcement Learning (RL) tasks. Just as toddlers evolve from exploration with sparse feedback to exploiting prior experiences for goal-directed learning with denser rewards, we simulate these analogous reward transitions in RL.  Inspired by this progression, we set out to explore the impact of various reward transitions on RL tasks. Our exploration revolves around the \textbf{Toddler-Inspired Reward Transition}. Through experiments in tasks like egocentric navigation and robotic arm manipulation, we found that reward transitions impact sample efficiency and success rates, with the toddler-inspired S2D transition being notably effective. Additionally, transitions, especially the S2D, smooth the policy loss landscape, promoting wider minima for better generalization. 

%-------------UPGRADE 버전 BY 준석
Toddlers evolve from free exploration with sparse feedback to exploiting prior experiences for goal-directed learning with denser rewards. Drawing inspiration from this \textbf{Toddler-Inspired Reward Transition}, we set out to explore the implications of varying reward transitions when incorporated into Reinforcement Learning (RL) tasks. Central to our inquiry is the transition from sparse to potential-based dense rewards, which share optimal strategies regardless of reward changes.
Through various experiments, including those in egocentric navigation and robotic arm manipulation tasks, we found that proper reward transitions significantly influence sample efficiency and success rates. Of particular note is the efficacy of the toddler-inspired Sparse-to-Dense (S2D) transition. Beyond these performance metrics, using \textit{Cross-Density Visualizer} technique, we observed that transitions, especially the S2D, smooth the policy loss landscape, promoting wide minima that enhance generalization in RL models.

%Through various experiments, including those in egocentric navigation and robotic arm manipulation tasks, we found that the toddler-inspired Sparse-to-Dense (S2D) transition significantly enhances sample efficiency and success rates.

\end{abstract}

\section{Introduction}

\begin{figure*}[t!]
\centering
\includegraphics[width=0.85\textwidth]{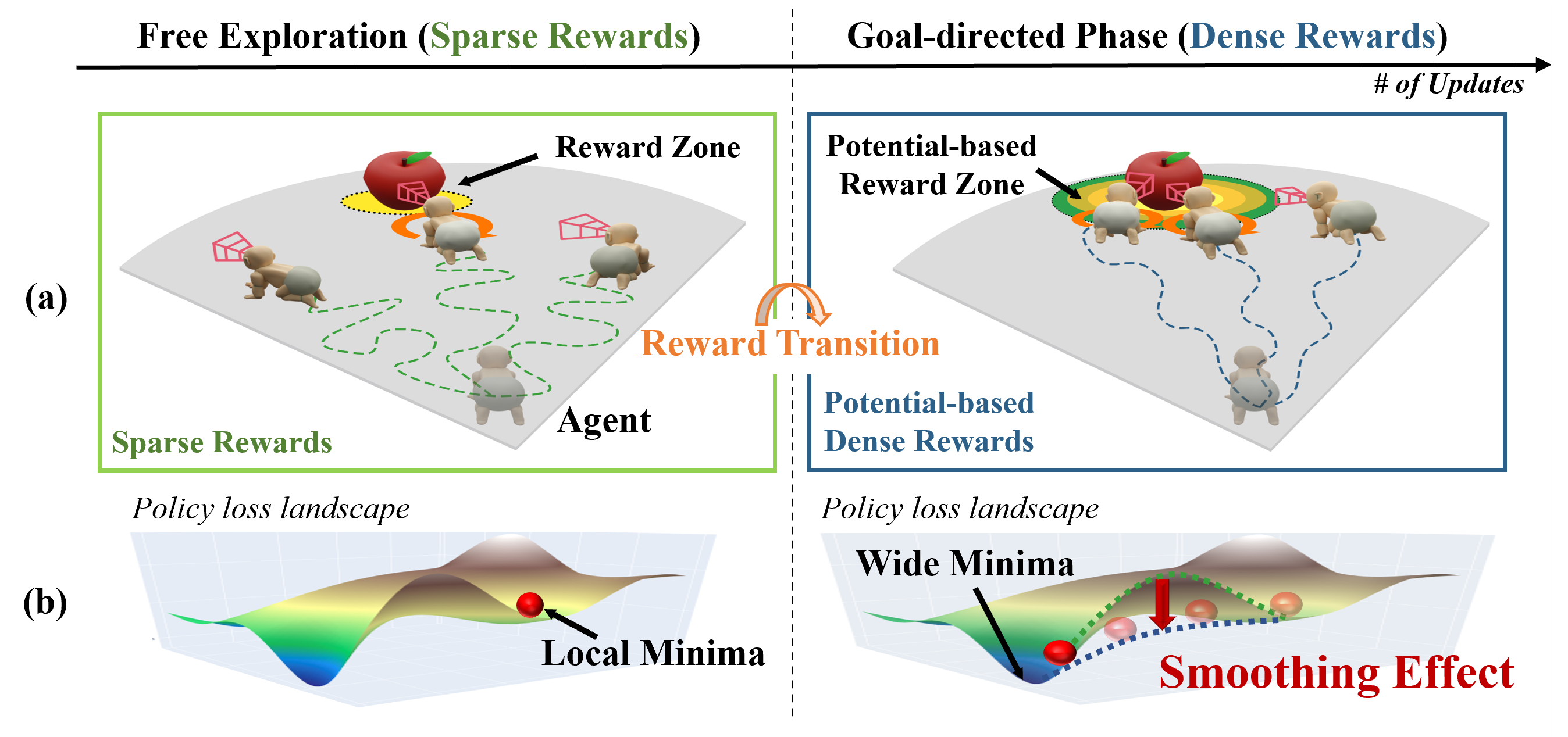}
\caption{Parallel learning trajectories: toddlers and agents. (a) The figure compares a toddler's learning journey with an agent's. On the left, a toddler freely explores the apple, symbolizing sparse reward learning. As we transition right, the toddler's focus on specific tasks reflects goal-guided learning. Similarly, the agent's progression from sparse to potential-based dense rewards is charted above, highlighting parallels in learning evolution. (b) As reward transitions occur, the depth of local minima reduces, leading to a wide minima via the smoothing effect, thereby enhancing more generalization.
}
\label{dummy1}
\end{figure*}

In early years, toddlers behave much like exploratory agents. Throughout their development, they interact with their surroundings without much prior knowledge, akin to someone embarking on new experiences without expecting immediate rewards~\cite{oudeyer2016evolution}. As they grow, toddlers transition from free exploration to more goal-directed learning, aiming for specific goals, resembling someone working towards known rewards for their efforts~\cite{gopnik1999scientist, Gibson, Piaget, gopnik2017changes} as illustrated in Figure~\ref{dummy1}.

This learning pattern in toddlers can be incorporated in Reinforcement Learning (RL), as illustrated in Figure~\ref{dummy1}-(a). In RL, agents learn by interacting with their environment and receiving feedback, much like how toddlers learn from their interactions. Similar to toddlers, agents must navigate towards positive feedback they receive, which can be infrequent (sparse) or detailed (dense). Sparse feedback might mean that the agent requires more attempts to figure out the desired behavior due to limited guidelines~\cite{andrychowicz2020learning,knox2023reward}. Meanwhile, dense feedback can guide the agent faster but might inadvertently focus them on immediate outcomes, missing out on the bigger picture or long-term strategies~\cite{laud2004theory}.

Given these intricacies, simply sticking to one type of feedback might not capture the essence of learning. Drawing inspiration from toddler developmental stages, blending both feedback types could provide richer insights. The transition toddlers make from free exploration -- akin to sparse feedback -- to specific, goal-driven learning -- similar to dense feedback in RL -- offers a unique perspective, as shown in Figure \ref{dummy1}-(a). With this perspective, our paper addresses the following question: \textit{``How does reward transition that proceeds in a sparse-to-dense manner, inspired by toddler learning, affect the learning of agents?''}
Through a series of experiments, including egocentric navigation and robotic arm manipulation tasks, we aim to explore the \textbf{Toddler-Inspired Sparse to Potential-based Dense (S2D) Reward Transition}. Our goal is not only to explore its efficacy but also to delve into the underpinning reasons by analyzing its comparative advantages against other rewards or prevalent strategies in the field of RL.

Taking the concept of ``reward transition in learning'' further, we consider visualizing the learning parameters as a topographical map. One type of such landscape representation, a policy loss landscape provides an intuitive visualization of learning dynamics in RL~\cite{li2018visualizing}. On this map, each point corresponds to a set of learning parameters, and its altitude signifies the loss value. As in any landscape, some areas are rugged, featuring steep mountains or deep valleys, indicating challenging learning regions.
Understanding the notion of smoothness of loss landscape is vital in neural network-based learning. Smooth terrains in this landscape, devoid of abrupt pitfalls, enable quicker and more reliable convergence through gradient descent.

Critically, smoother landscapes often promote wide minima, which are associated with better generalization of learned policies to novel situations in dynamic environments~\cite{keskar2017on}.
Empirically, we observed that employing the Sparse-to-Dense (S2D) Reward Transition made this terrain smoother by reducing the depth of local minima, as illustrated in Figure~\ref{dummy1}-(b). 

Our study contributes to a deeper understanding of the intricate balance between exploration and exploitation and provides insight into designing reward structures in RL. By emulating the learning processes observed in toddlers, we hope to bridge the gap between biological and artificial learning mechanisms. Based on this, we have offered a novel perspective in applying and addressing this synthesis for more robust, adaptable, and efficient RL systems.

The main contributions of this paper can be summarized as follows: \textbf{(1)} We observed that the Toddler-Inspired Reward Transition enhances success rates, sample efficiency, and generalization within goal-conditioned RL. \textbf{(2)} Our experimental analyses support that such transition has smoothing effects on the policy loss landscape and promotes wide minima, corroborating the performance improvements. \textbf{(3)} Our findings highlight the potential of biologically inspired approaches in providing clues for exploration-exploitation tradeoff and reward shaping challenges.

% Rough terrains, in contrast, can trap algorithms in local minima or cause oscillations, impeding learning.
%This progression from sparse to dense reward interactions is not just coincidental; it is a testament to the significance of reward transitions in learning.
%A reward transition during these formative years is paramount. If misaligned or inadequate, the toddler could face challenges, potentially struggling or failing to acquire certain skills~\cite{nickersonCP}.
% Smoother terrains like open plains suggest regions where learning is more stable and effective.
% A smoother landscape implies that our reward framework can more easily determine the best actions or decisions, minimizing confusion or missteps. 
\section{Related Work}

\subsubsection{Toddler-inspired learning.}
Drawing insights from toddler developmental stages has provided a fresh perspective in advancing deep learning. By harnessing the innate exploratory tendencies and distinctive learning mechanisms of toddlers, researchers have refined both supervised and reinforcement learning techniques.
% Emulating toddlers' explorative learning has optimized supervised and reinforcement techniques.
% For instance, toddlers' varied object exploration inspired datasets processed by CNNs, which then outperformed those built on adult perspectives~\cite{bambach2018toddler}.
For instance, classifiers trained on datasets of toddlers' perspectives on objects outperformed those using adults' perspectives~\cite{bambach2018toddler}, underscoring the potential of leveraging toddlers' exploration mechanism.
Similarly, critical learning phases in toddlers have counterparts in RL~\cite{park2021toddler, de2022critical} and deep networks~\cite{achille2018critical}.  Such toddler-inspired methodologies emphasize the alignment between toddler growth and AI model evolution, underscoring the potential of biological insights in driving AI forward.

\subsubsection{Exploration-exploitation in deep RL.}
% RL agents grapple with exploration versus exploitation. Balancing these is key to RL efficiency~\cite{ladosz2022exploration}.
Balancing exploration with exploitation is an inherent challenge of RL~\cite{ladosz2022exploration}. Moreover, deep RL intensifies this complexity with high-dimensional spaces, like raw image inputs, where pixels are unique dimensions. To mitigate this, many algorithms favor exploration, utilizing tools like intrinsic motivation~\cite{badia2020never,aubret2019survey,pathak2017curiosity}. Drawing from toddlers' learning, we observe the transition from free exploration to specific exploitation based on collected previous experience. Our approach offers a fresh take on RL's exploration-exploitation dilemma without introducing algorithmic complexities.

\subsubsection{Curriculum learning.} Curriculum Learning (CL), inspired by curricula in human's education, has been known to boost performance, training speed~\cite{Hacohen2019OnTP}, and safety~\cite{Turchetta2020SafeRL} in machine learning. CL's progression from simple to complex tasks promotes generalization and convergence~\cite{Bengio2009CurriculumL, Weinshall2018CurriculumLB} in both supervised and reinforcement learning~\cite{florensa2018automatic, graves2017automated, narvekar2020generalizing}.
Unlike the studies that initially restrict the diversity to easy tasks~\cite{kalAntidis20hard, du21self, dong2017class}, several studies~\cite{zhang1994selecting, mackay1992information} advocate a \textit{general-to-specific} approach, where agent initially collects varied learning experiences and then exploits these experiences later in the curriculum. Adapting the S2D transition observed in toddlers into RL, we embrace this philosophy in reward transition for goal-oriented tasks.

\subsubsection{Potential-based reward shaping (PBRS).}
In RL, maximizing cumulative rewards guides agent behavior. However, due to the inherent challenges in designing optimal reward functions for various tasks, it often necessitates a process known as reward engineering. Reward Shaping (RS) is one such technique that enhances training by supplementing the environment's feedback ~\cite{taylor2009transfer}. Particularly, in environments where the reward changes, an additional reward from a \textit{potential function} is employed to ensure the agent's optimal strategy remains unaffected~\cite{ng1999policy}. Commonly, such shaped rewards are consistently applied throughout training. Unlike this, we explore the \textbf{Toddler-Inspired Reward Transition}, focusing on the effects of changing the density of rewards. %. Instead of sticking to a one-size-fits-all reward approach, we focus on reward structure that mirrors how toddlers learn. 

\begin{figure}[t!]
    \centering
    \includegraphics[width=0.9\columnwidth]{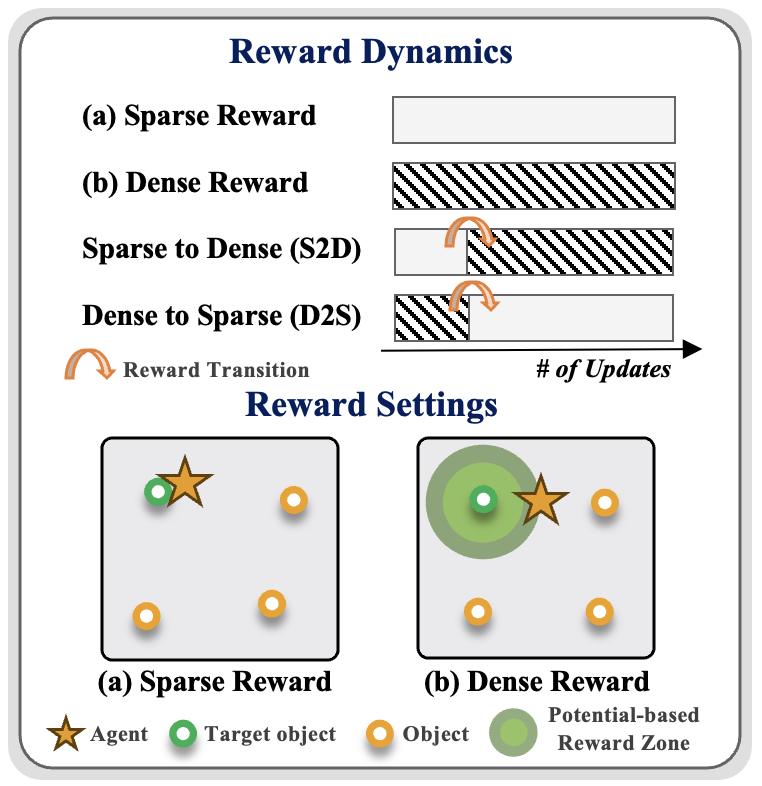}
    %\vskip -0.1in
    \caption{Overview of the baseline rewards. The S2D presents reward inspired by toddler learning. In sparse rewards, agents are rewarded upon reaching the target. For potential-based dense rewards, they get an extra reward determined by the distance to a specific unit from the object. }%Initially, sparse rewards encourage free exploration, later transitioning to dense rewards using PBRS. This progression is contrasted with the vice versa approach (D2S), solely using either sparse or dense rewards, and algorithmic solutions grounded in intrinsic motivation. Together, they offer a comprehensive comparison of various exploration-exploitation strategies of environments. }

    \label{figure:losslandscape}
\end{figure}

\begin{comment}
\begin{figure*}[t!]
\centering
\includegraphics[width=0.9\columnwidth]    {icml2023/picture/finalmain2.png} % 
\vskip -0.1in
\caption{Overview of the baseline rewards. The S2D presents reward inspired by toddler learning. Initially, sparse rewards encourage free exploration, later transitioning to dense rewards using PBRS. This progression is contrasted with the vice versa approach (D2S), solely using either sparse or dense rewards, and algorithmic solutions grounded in intrinsic motivation. Together, they offer a comprehensive comparison of various exploration-exploitation strategies of environments.}
\label{Schematic overview}
\vskip -0.2in
\end{figure*}
\end{comment}

\section{Preliminaries}
\subsubsection{Reinforcement learning.} RL is a field of machine learning in which the agent learns through trial and error, similar to how humans acquire skills.
It is applied to various tasks that involve sequential decision making. A widely used formulation of RL problem, Markov Decision Process (MDP) is defined as $\langle \mathcal{S}, \mathcal{A}, \mathcal{P}, \mathcal{R}, \gamma\rangle$, where
$\mathcal{S}$ is a set of environment states, $\mathcal{A}$ is a set of possible actions, $\mathcal{P}: \mathcal{S} \times \mathcal{A} \rightarrow  \Delta(\mathcal{S})$ is a transition probability distribution, $\mathcal{R}: \mathcal{S}\times \mathcal{A} \rightarrow \mathbb{R}$ is a reward function and $\gamma$ is a discount factor.
At every time step $t\in \mathbb{N}$, the agent in the current state $s_t \in \mathcal{S}$ performs the action $a_t \in \mathcal{A}$ according to a policy  $\pi(\cdot|s_t)$, and receives the next state $s_{t+1} \sim \mathcal{P}(\cdot|s_t,a_t)$ and reward $\mathcal{R}(s_t,a_t)$. RL algorithms aim to obtain an optimal policy $\pi^*\in \Pi^*$ that maximizes the expected cumulative rewards $R = \mathbb{E}\left[\sum_{t=0}^{\infty} \gamma^{t}\mathcal{R}\left(s_{t}, a_{t}\right)\right]$
% \begin{equation}
% \label{eq:cumulative_rewards}
% R_{\mathcal{M}, \pi} = \mathbb{E}_{a_t\sim\pi(\cdot|s_t), s_{t+1}\sim\mathcal{P}(\cdot|s_t, a_t)}\left[\sum_{t=0}^{\infty} \gamma^{t}\mathcal{R}\left(s_{t}, a_{t}\right)\right]    
% \end{equation}
with $\gamma$ applied, where $\Pi^*$ is the set of optimal policies. 

\subsubsection{Curriculum learning.}
\label{sec:currl}
Curriculum Learning (CL) is a strategy to train an ML model using tasks that gradually increase in difficulty. In RL, CL can be formulated as a learning framework where the agent is trained on a sequentially changing series of MDPs $\mathcal{M}_i = \langle \mathcal{S}, \mathcal{A}, \mathcal{P}, \mathcal{R}_i, \gamma \rangle$.

\begin{definition}[Curriculum]
\label{def:curriculum}
Let $\mathcal{M}_1, \mathcal{M}_2, \cdots, \mathcal{M}_N$ be a sequence of MDPs $\mathcal{M}_i = \langle \mathcal{S}, \mathcal{A}, \mathcal{P}, \mathcal{R}_i, \gamma \rangle$, and $\mathcal{T} = (T_1, T_2,\cdots,T_{N-1}) \in \mathbb{N}^{N-1}$. A \textit{curriculum} is a tuple $\mathscr{C} = (\{\mathcal{M}_i\}_{i=1}^N, \mathcal{T})$ where the agent is trained on $\mathcal{M}_{I(t;\mathcal{T})}$ at training step $t$. 
Stage indicator $I(t;\mathcal{T})$ is defined as:
\[
    I(t;\mathcal{T}) := i,\quad t \in \left[T_{i-1}, T_i\right)
\]
for each stage $i \in \{1, \cdots, N\}$, where $T_0:=0$ and $T_N:=\infty$.
We call $\mathcal{T} = (T_1, T_2, \cdots, T_{N-1})$ the \textit{stage transitions}.
\end{definition}

CL also boosts the training by arranging the tasks as \textit{``general-to-specific,''} where the agent is provided rewards with monotonically increasing densities over training.
Formally, we say an environment is sparse when only a small portion of the state space is included in $\text{supp}(\mathcal{R})$. That is,
\begin{align*}
&|\text{supp}(\mathcal{R})| \ll |\mathcal{S}|, \,\,\, \text{where}\\
\text{supp}(\mathcal{R})&=\{s\in \mathcal{S} \mid \exists a\in \mathcal{A} \,\,\, s.t. \,\,\, \mathcal{R}(s, a) \neq 0 \}.
\end{align*}
$\text{supp}(\mathcal{R})$ means the region \textit{supported} by reward function $\mathcal{R}$ which has non-zero reward for some actions.

%\subsubsection{Wide Minima Phenomenon.}
%It's vital to note that during the training and testing phases, the data distributions may differ—this phenomenon is known as the data distribution shift. Imagine a neural network's training journey as navigating through hilly terrain, called the loss landscape~\cite{li2018visualizing}. Here, we aim to find low points or 'minima'. Wide and flat regions, known as 'wide minima', lead to more stable learning and better generalization with new data~\cite{keskar2016large}. In contrast, sharp and pointed areas, or 'sharp minima', can make models overly specific and less generalizable with new data~\cite{goodfellow2014qualitatively}. Research shows that aiming for wide minima often results in better performance~\cite{keskar2017on, j2018finding}. In deep RL, a sequence of data distribution the agent experiences, each data will follow a slightly changed distribution from each other. Therefore, a network in wide minima will improve a generalization ability.

%--------준석 improve----
\subsubsection{Wide minima phenomenon and loss landscape.}

Deep neural networks traverse a high-dimensional \textit{loss landscape}, with altitude indicating the loss for specific parameters~\cite{li2018visualizing}. The aim is to find the \textit{minima}. In \textit{wide minima}, due to broad gradients, gradient descent is more likely to converge smoothly to global minima. This fosters robustness and superior generalization to new data~\cite{keskar2016large}. Conversely, in \textit{sharp minima}, steep gradients can trap models in local minima, resulting in overfitting and poor generalization across diverse data distributions~\cite{goodfellow2014qualitatively}. Empirically, models within wide minima demonstrate better performance and generalization than those in sharp minima~\cite{keskar2017on, j2018finding}. In deep RL as well, where the distribution of agent's experiences may slightly vary every time step, policies in wide minima could improve in generalization.

\section{\textit{Toddler-Inspired} Reward Transition} To conduct our experiments, we need to formulate and design a toddler-inspired reward transition in RL, emulating the toddler reward transition paradigm. Furthermore, we analyze the influence of this reward transition on the learning behavior of agents, focusing on the policy loss landscape and the wide minima phenomenon.

\subsection{Toddler-Inspired Sparse to Potential-based Dense Reward Curriculum}

In this paper, we harness curriculum learning from the perspective of encouraging exploration-to-exploitation as a Sparse to potential-based Dense (S2D) reward transition curriculum and explain how this learning mechanism can benefit RL. We define $\mathscr{C} = (\{\mathcal{M}_i\}_{i=1}^N, \mathcal{T})$ as an \textit{S2D-curriculum} if reward functions of MDPs $(\mathcal{M}_1, \mathcal{M}_2,\cdots,\mathcal{M}_N)$ become progressively denser while preserving optimality.

\begin{definition}[\textit{Toddler-inspired S2D-curriculum}]
\label{def:Anti_curriculum}
A curriculum $\mathscr{C} = (\{\mathcal{M}_i\}_{i=1}^N, \mathcal{T})$ with MDPs $\{\mathcal{M}_i\}_{i=1}^N$  is an \textit{S2D-curriculum} if following conditions are satisfied:%$\text{supp}(\mathcal{R}_1)\subseteq \text{supp}(\mathcal{R}_2) \subseteq \cdots \subseteq \text{supp}(\mathcal{R}_N)$ and $\Pi^*_1\supseteq \Pi^*_2 \supseteq \cdots \supseteq \Pi^*_N$ hold, where $\Pi^*$ is the set of optimal policy that maximizes the expected cumulative rewards.%
\begin{equation}
\mathrm{supp}(\mathcal{R}_1)\subseteq \mathrm{supp}(\mathcal{R}_2) \subseteq \cdots \subseteq \mathrm{supp}(\mathcal{R}_N) 
\label{cond1}
\end{equation}
\begin{equation}
\Pi^*_1\supseteq \Pi^*_2 \supseteq \cdots \supseteq \Pi^*_N, 
\label{cond2}
\end{equation}
$\Pi_{i}^*$ is a set of optimal policies with MDP $\mathcal{M}_i$. Here, we call $\{\mathcal{R}_i\}_{i=1}^{N}$ a \textit{guidance} of the curriculum $\mathscr{C}$.
\end{definition}

The first condition in Equation \ref{cond1}
denotes that the reward function becomes denser, \emph{i.e.}, the guidance becomes more explicit.
The second condition in Equation \ref{cond2}
constrains the optimality to be preserved during the transition of the MDPs, \emph{i.e.}, the optimal policies of $\mathcal{M}_i$ are also optimal in $\mathcal{M}_{i+1}$.
At a high level, the sequence of MDPs in the above definition is \textit{S2D-curricular} in the sense that the reward functions are arranged in a \textit{``sparse-to-dense''} order.

%\section{Methodology for Analyses}

\subsection{Visualizing Post-Transition 3D Policy Loss Landscape: Cross-Density Visualizer}
\label{visual:cross_density_visualizer}

As seen in Figure~\ref{figure:visualpolicylossland} and Appendix B, our study delves into the effect of the S2D transition on policy loss landscape, reflecting toddlers' cognitive evolution~\cite{gopnik2017changes, Piaget}. Following \cite{li2018visualizing}, we visualize the policy loss landscapes using grids of parameters $\tilde{\theta} = \theta + \alpha \mathbf{x} + \beta \mathbf{y}$. Here, $\theta$ signifies current parameters and $\alpha$, $\beta$ are normalized coordinates. Vectors $\mathbf{x}$ and $\mathbf{y}$ that form the axes arise from two specific perturbations in the network's parameter space. $\mathbf{x}$ and $\mathbf{y}$ are made unit vectors by a normalized filter for consistent scaling and clarity. The Z-axis captures policy loss, averaged over a batch of transitions from the replay buffer. We are not concerned with the altitude and which landscape is above or below another, because the two landscapes correspond to separate network parameters, each of them having its own loss range according to its current learning progress.

Noting the lack of visualization techniques in prior studies for policy loss landscapes based on reward transitions, we design the Cross-Density Visualizer. This method portrays the 3D policy loss landscape during transitions from purely sparse or dense rewards to mixed-reward settings. Thus, one set includes Sparse-to-Dense (S2D) and Sparse-to-Sparse (Only Sparse), while the other contains Dense-to-Sparse (D2S) and Dense-to-Dense (Only Dense). Our representations, displayed in Figure~\ref{figure:visualpolicylossland} and expanded upon in Appendix B, highlight comparable \textit{smoothing effects} particularly in the S2D model. %, distinguishing it from the rest.

%\subsubsection{Methodology for visualizing the policy loss landscape.}

%It should be noted that we are not concerned with the altitude and which landscape is above or below another because the landscape altitude can differ due to the neural networks' adaptability. 
%\subsubsection{Methodology for Visualizing the Policy Loss Landscape}
%We follow the scheme from \cite{li2018visualizing} for visualizing the policy loss landscapes. We use $\tilde{\theta} = \theta + \alpha u + \beta v$ where $\theta$ is the current set of parameters, and normalized $\alpha, \beta$ are coordinates in x, y-axes. The X and Y-axes represent random orthogonal gradient directions, with the normalized Z-axis denoting policy loss.  The policy loss is calculated by averaging the policy losses, calculated according to the parameter $\tilde{\theta}$, across a batch of transitions from the replay buffer. The heights of the loss landscapes vary, since the neural networks can flexibly approximate the same loss function via different parameters. Also, the height order within each pair of loss landscapes within every subfigure may change, since the reward from the potential-based dense guidance is neither strictly positive nor strictly negative.

\subsection{Exploring Minima Sharpness After Reward Transitions}
Observing a reduction in the depth of local minima due to smoothing effects led us to hypothesize that the S2D transition promotes escape from local minima and enhances generalization in wide minima. Wide minima in neural networks can serve as a measure indicating robust and adaptable models~\cite{keskar2017on, j2018finding}. By exploring minima with this transition, we aspire for both performance and a deeper grasp of agent adaptability in diverse scenarios. To check how much the policy resides in a wide minima, we measure the end-of-training convergence of S2D's neural network to wide minima using the sharpness metric of Equation~\ref{defn:sharpness} and compare with those of baselines in the same way as proposed in \cite{foret2021sharpnessaware}, which is a specific form of sharpness measure proposed in \cite{keskar2017on}. 
\begin{equation}
    \max_{||\epsilon||_2\leq \rho} L_{\pi}(\theta+\epsilon)-L_{\pi}(\theta)
    \label{defn:sharpness}
\end{equation}
Here, $\theta$ is the current parameter in the policy loss landscape.
The maximizer $\epsilon$ can be estimated from
$
\hat{\epsilon}=\rho \,\text{sgn}(\nabla_\theta L_{\pi}(\theta))\cdot \nicefrac{|\nabla_\theta L_{\pi}(\theta)|^{q-1}}{\big( ||\nabla_\theta L_{\pi}(\theta)||^q_q\big)^{\frac{1}{p}}},
$
where $1/p + 1/q = 1$, $\text{sgn}
(\cdot)$ is element-wise sign function~\cite{foret2021sharpnessaware}.
We use $\rho=0.02, p=2$ in our experiments.

% where $\text{sgn}(\cdot)$ is element-wise sign function, and we use $p=q=2$ and $\rho=0.02$ in our experiments.
% \footnote{This is the closed-form solution to the dual norm problem, yielded by applying Taylor expansion to the optimization problem within $\rho$-ball \cite{foret2021sharpnessaware}.}
% Measurement of sharpness can be calculated as the difference between maximum loss near radius $\rho \text{-ball}=\{||\epsilon||_2 \leq \rho\}$ and current loss. 
%In the same way, the maximizer $\epsilon$ can be estimated from

\begin{figure}[t]
\centering
    \subfloat[][ViZDoom-Seen]{ \label{env:vizdoom-seen}
      \includegraphics[width=0.48\columnwidth]{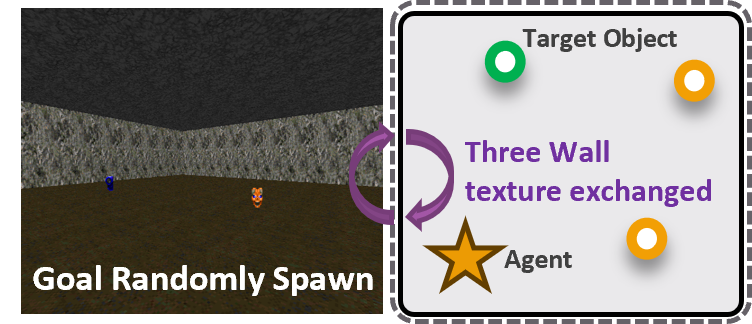}
    }
    \subfloat[][ViZDoom-Unseen]{ \label{env:vizdoom-unseen}
     \includegraphics[width=0.48\columnwidth]{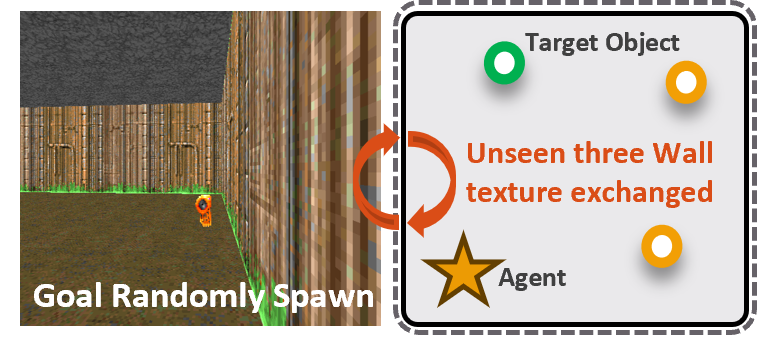}
    }\\
    \subfloat[][LunarLander-V2]{ \label{evn:lunar}
      \includegraphics[width=0.31\columnwidth]{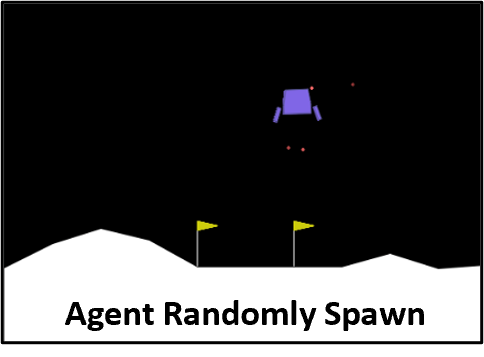}
    }
    \subfloat[][CartPole-Reacher]{ \label{env:cartpole}
      \includegraphics[width=0.31\columnwidth]{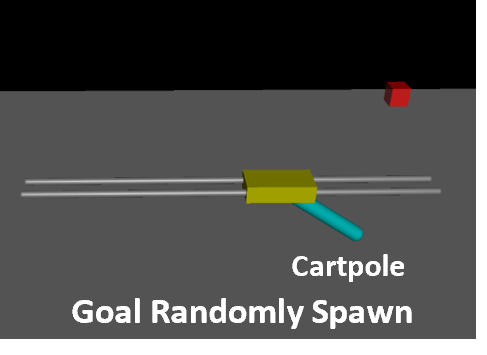}
    }
    \subfloat[][UR5-Reacher]{ \label{env:ur5}
     \includegraphics[width=0.31\columnwidth]{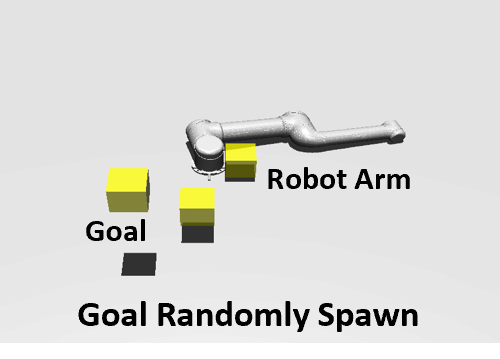}
    }
    \caption{Experimental environments. In particular, (a) and (b) are environments for evaluating generalization. %(c), (d) and (e) are environments for continuous action space.
    }
    \label{allenvvv} % allEnvresult
\end{figure}

\begin{figure*}[t]
\centering
    \subfloat[][CartPole-Reacher]{ \label{exp:cartpole}
     \includegraphics[width=0.3133\textwidth]{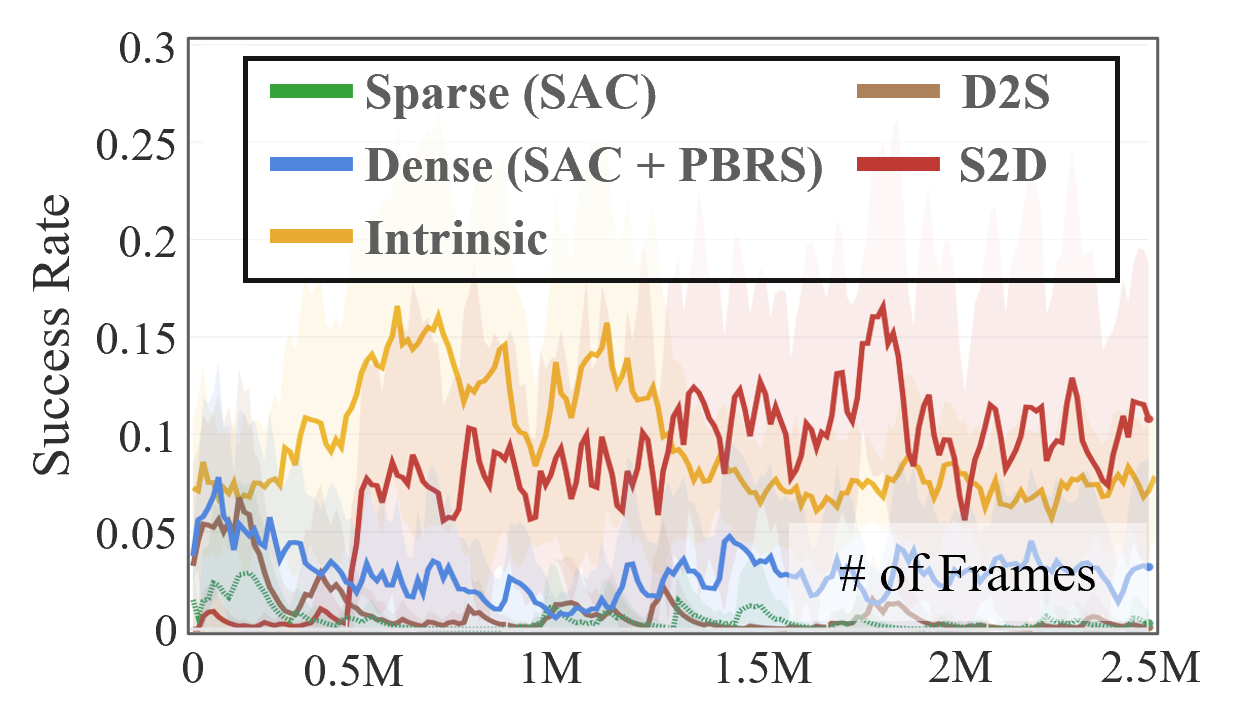}
    }
    \subfloat[][UR5-Reacher]{ \label{exp:ur5}
      \includegraphics[width=0.3133\textwidth]{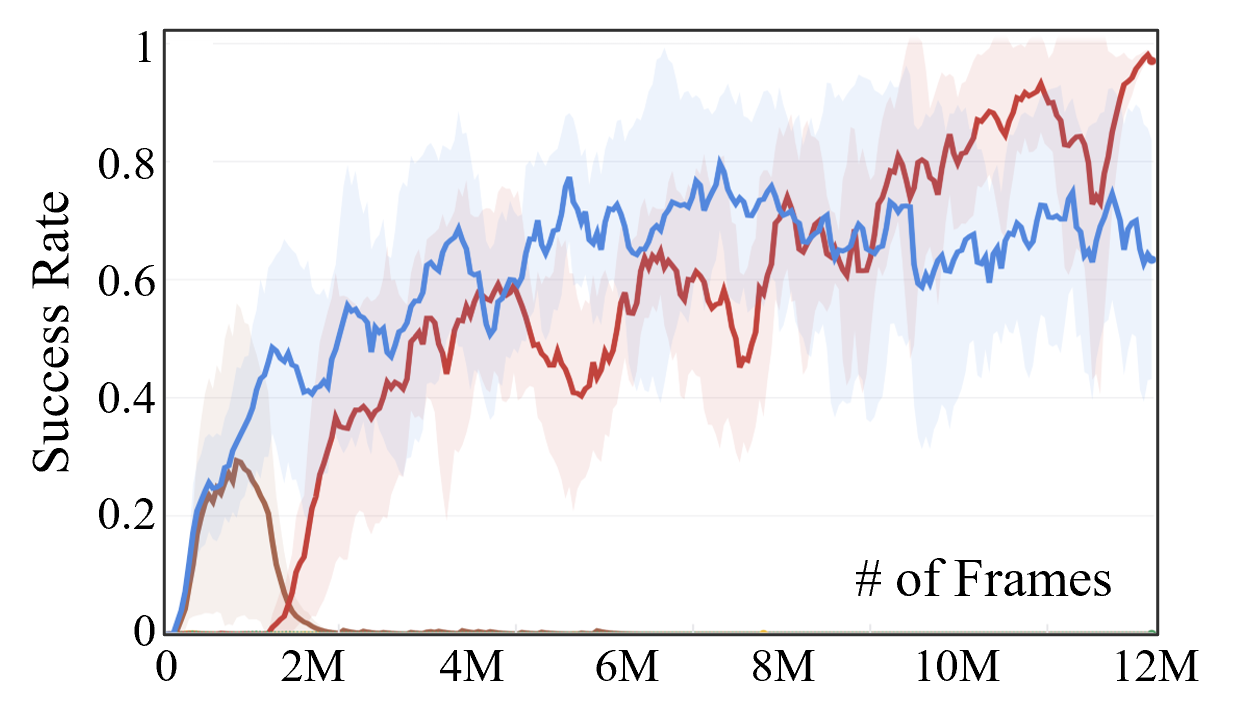}
    }
    \subfloat[][LunarLander-V2]{ \label{exp:lunar}
      \includegraphics[width=0.3133\textwidth]{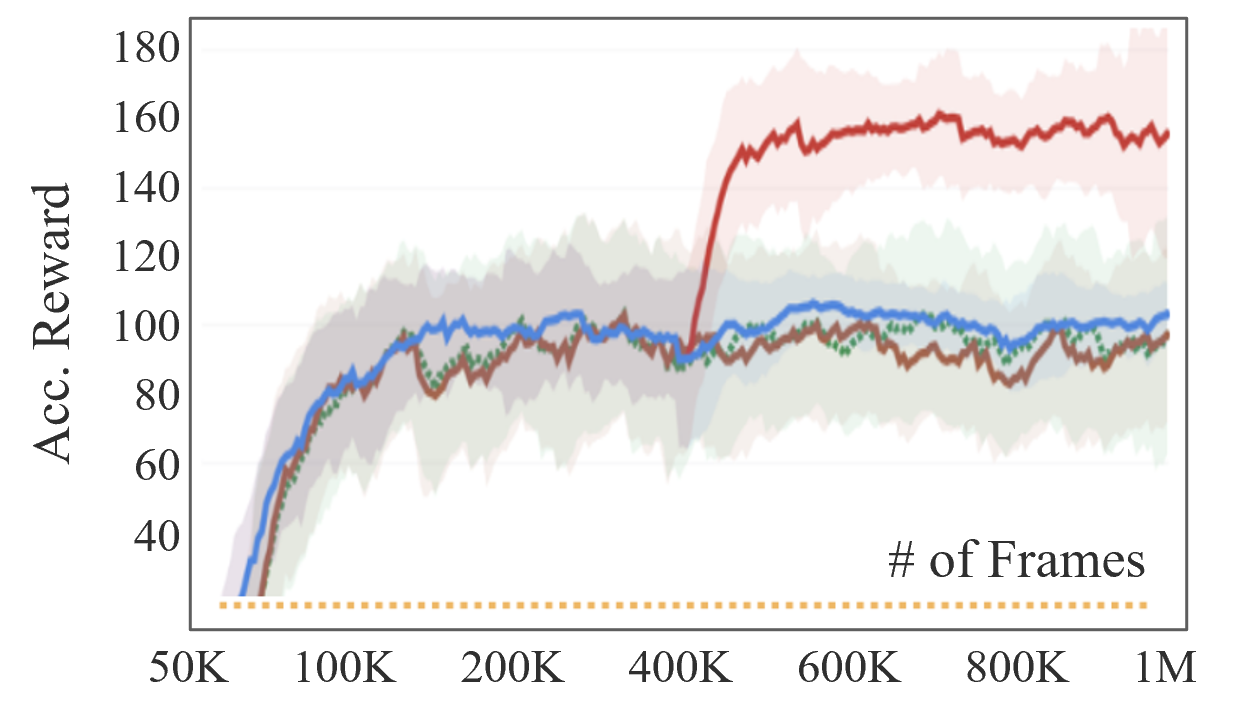}
    }
    \caption{Performance of the agent with various reward types in multiple goal-oriented tasks. Notably, in (c) LunarLander, the accumulated reward from intrinsic rewards was significantly below zero, indicated by a dashed line.
    % For Cartpole-Reacher and UR5-Reacher, the y-axis shows the success rate. The x-axis, across all environments, indicates the number of updates. In LunarLander, the y-axis uses the commonly measured accumulated reward.
    }
    \label{fig:main_exp} % allEnvresult
\end{figure*}

\begin{figure*}[t]
\centering
    \subfloat[][ViZDoom-Seen]{ \label{exp:vizdoom-seen}
      \includegraphics[width=0.47\textwidth]{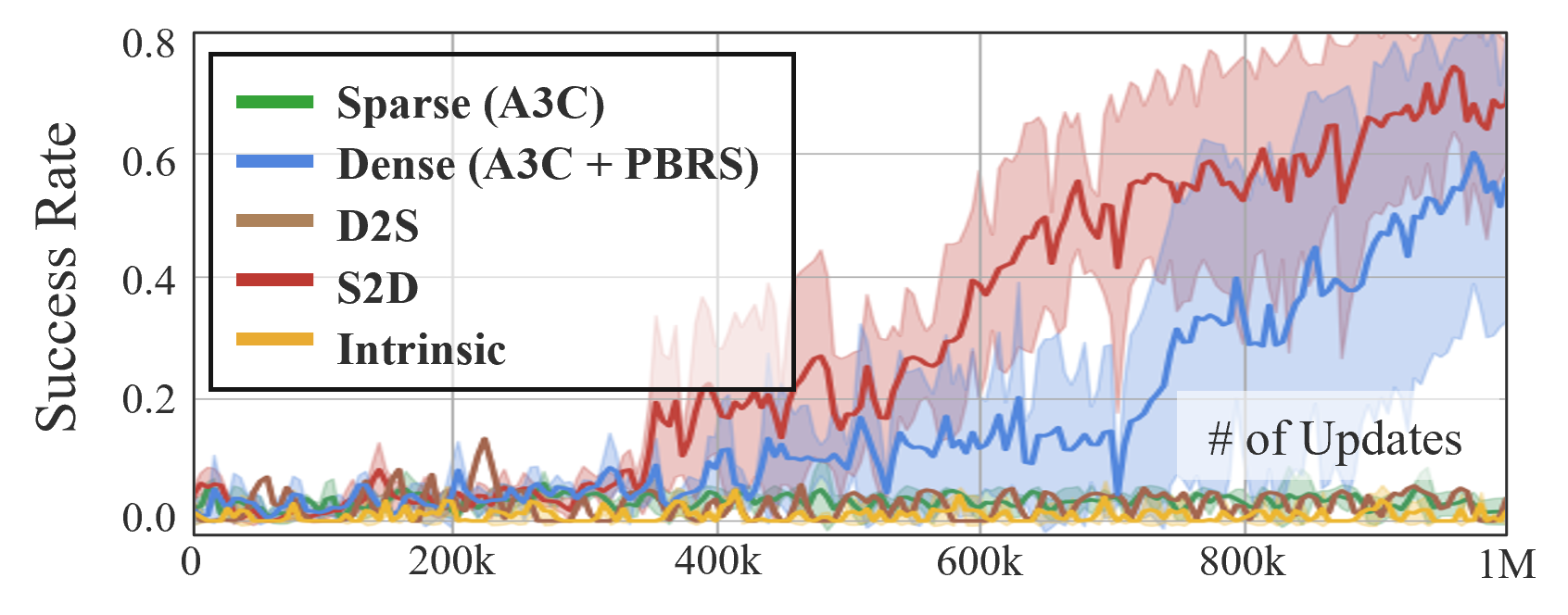}
    }
    \subfloat[][ViZDoom-Unseen]{ \label{exp:vizdoom-unseen}
     \includegraphics[width=0.47\textwidth]{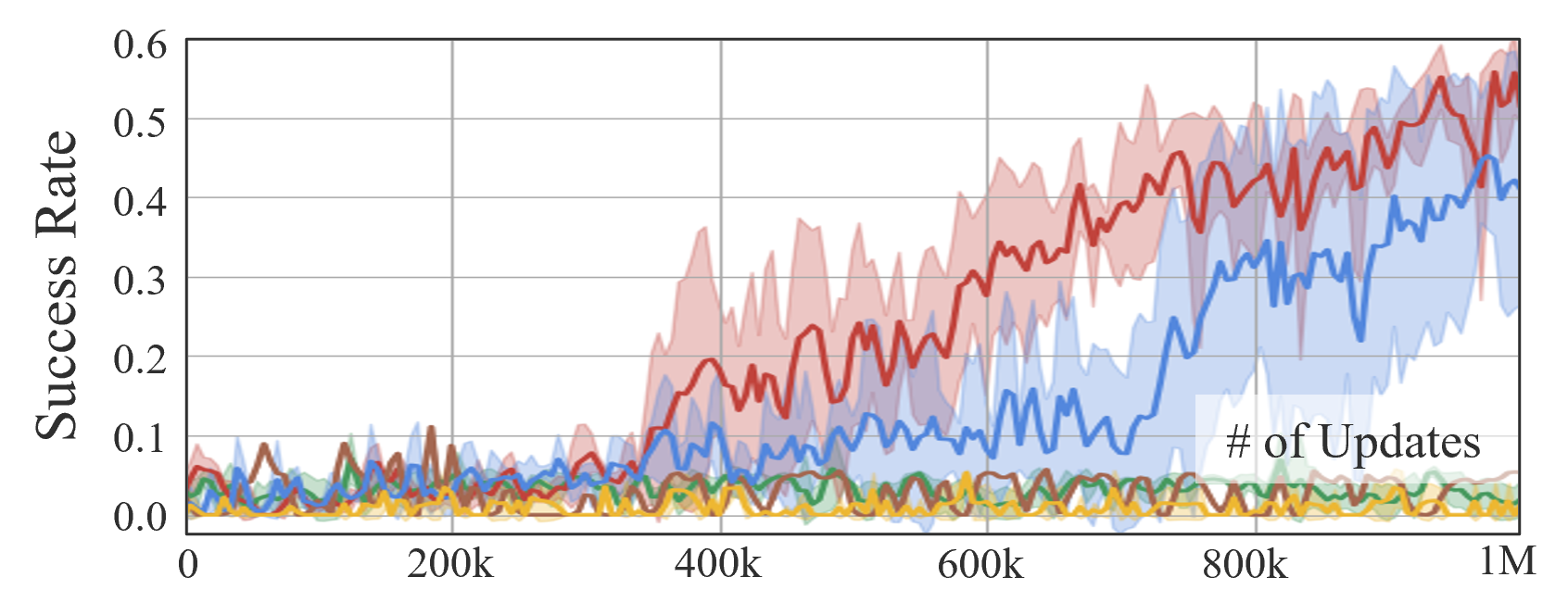}
    }
    \caption{ Generalization performance of the ViZDoom agent with various types of rewards.
% The y-axis represents the success rate, while the x-axis indicates the number of updates.
}
    \label{fig:vizdoom} % allEnvresult
\end{figure*}

\begin{table*}[t]
    \centering
    \vspace{0.25em}
    \begin{adjustbox}{width=1\textwidth}
    \begin{tabular}{cccccccccc}
    \toprule
Task  & Metric & S2D($\mathscr{C}_1$)    & S2D($\mathscr{C}_2$)     & \textbf{S2D}($\mathscr{C}_3$)   & Only Sparse & Only dense & D2S($\mathscr{C}_1$) & D2S($\mathscr{C}_2$) & D2S($\mathscr{C}_3$)  \\
\midrule
\multirow{2}{*}[0.3ex]{\makebox[0pt]{\LARGE \thead{Lunar\\Lander}}}   & Perf. & 138.71\stdv{3.71}          & 63.40\stdv{160.55}   & \textbf{168.88}\stdv{23.66}   & 142.50\stdv{4.25}          & 139.68\stdv{14.90}          &    140.75\stdv{7.46}     & 130.63\stdv{19.69}  &   142.373\stdv{15.62}\\
&  Sharp. & 27.06\stdv{36.31}  & 1231.93\stdv{2424.61}     & \textbf{7.46}\stdv{3.37} & 8.97\stdv{2.83}  & 8.71\stdv{4.43}   &   8.95\stdv{2.89}  & 8.99\stdv{2.97}& 11.32\stdv{3.72}\\
\midrule
\multirow{2}{*}{CartPole}    &    Perf.    & 3.18\stdv{4.0} & \textbf{14.61}\stdv{10.96}   & 5.29\stdv{7.472}  & 0.14\stdv{0.25}  & 3.88\stdv{4.63}  & 1.55\stdv{0.29} & 0.38\stdv{0.07} & 0.97\stdv{0.19}\\
&  Sharp. & 0.12\stdv{0.24}          & \textbf{0.01\stdv{0.15}}          & 0.01\stdv{0.24}          & 0.08\stdv{0.57}        & 0.19\stdv{0.03}          &  0.16\stdv{0.09} & 0.05\stdv{0.21} & 0.02\stdv{0.17} \\
\midrule
\multirow{2}{*}{UR5}    & Perf.    & 65.54\stdv{10.86}  & 65.69\stdv{17.32}      & \textbf{94.15}\stdv{4.28}      & 0.00\stdv{0.00}          & 64.23\stdv{13.03}          & 0.00\stdv{0.00}   & 0.00\stdv{0.00} & 0.00\stdv{0.00} \\
& Sharp. & 0.67\stdv{0.01}          & 0.62\stdv{0.11}          & \textbf{0.61\stdv{0.04}}          & 0.09\stdv{0.52}          & 0.67\stdv{0.01}          & 0.52\stdv{0.24}  & 0.56\stdv{0.28} & 0.47\stdv{0.20} \\
\bottomrule
\end{tabular}
\end{adjustbox}
\caption{Performance and sharpness metric measured for more than 5 different random seeds in each environment. We highlight the best performance and its sharpness values in bold, confirming that the top-performing \textbf{S2D} also resides in the widest minima.}
\label{table:sharpness}
\end{table*}

\section{Experiments}
%Env
In our experimental section, we delve deeply into the dynamics of the S2D reward transition compared to multiple reward-driven methods. % across various RL agents.
We unveil its substantial effects within multiple challenging environments, which are illustrated in Appendix A. We particularly explored the implications of applying the reward transition to RL by probing three critical questions:
\begin{tcolorbox}[width=\columnwidth,colback=white]
    \begin{itemize}[leftmargin=0pt]
    \item \textbf{Performance Enhancement}:  How does the Toddler-inspired S2D reward transition compare to other diverse reward settings?
    \item \textbf{Post-Transition 3D Policy Loss Landscape}: What are the effects of the S2D transition on the policy loss landscape?
    \item \textbf{Correlation Between Wide Minima and Toddler-Inspired Reward Transition}: Does the S2D transition foster convergence to wide minima?
    \end{itemize}
\end{tcolorbox}

Our understanding is further supported by extensive supplementary experiments in Appendix C.

\subsection{Reward Setting Details}
\subsubsection{Design of sparse and dense reward.}
In the sparse reward setting, the agent receives a reward only upon success, i.e., when reaching the goal. In the dense reward setting, the agent receives a potential-based reward based on its proximity to the goal. This is expressed as $\psi(s) = \diam(\mathcal{S}) - ||s - g||_2$, where $\mathcal{S}$ is the set of states, $g \in \mathcal{S}$ is the goal state, and $\diam(\cdot)$ denotes the diameter of the given set. %and $\mathcal{G}$ are states and goals respectively, and $\diam(\cdot)$ denotes the diameter of the given set.

%—from initial free exploration with sparse rewards to goal-guided learning—
\subsubsection{Reward-driven baselines for comparison.} 
Aiming to investigate the relation between exploration-exploitation tradeoff and reward transitions, we explore the Sparse-to-Dense (S2D) approach to mirror toddlers' developmental progression. We also evaluate its counterpart, Dense-to-Sparse (D2S), and solely sparse or dense reward schemes for a comprehensive assessment of reward strategies. Given the prominence of intrinsic motivation in tackling exploration-exploitation, we adopted NGU~\cite{badia2020never}—designed for discrete environments like ViZDoom and LunarLander—and RND~\cite{burda2018exploration}—tailored for continuous action such as CartPole and UR5—as additional baselines. 
Both methods incentivize agents to explore by providing additional  intrinsic rewards for discovering novel states.

%\subsubsection{Reward transition.} We also explored the best timing hyperparameter for reward transition through ablation studies as seen in Table~\cite{table:sharpness}. Heeding the importance of the temporal aspects of initial cognitive and motor interactions in early developmental stages~\cite{piaget1952origins, shonkoff2000neurons}, we divided our entire training duration using the first quarter as a reference and segmented it into three points. These transitions occur at \( t \in \{1N, 2N, 3N\} \), where \( N \) represents roughly a third of the first quarter of the entire training duration. Specific settings for \( N \), tailored to each environment's episode length, can be found in \cref{table:hyper2} of the Appendix. We denote these phases as \( \mathscr{C}_1 \), \( \mathscr{C}_2 \), and \( \mathscr{C}_3 \), signifying the S2D or D2S reward transition points.

\subsubsection{Design for reward transition.} We also explored the hyperparameter for reward transition through ablation studies as seen in Table~\ref{table:sharpness}. Recognizing the importance of the temporal aspects of initial cognitive and motor interactions in early developmental stages~\cite{Piaget,shonkoff2000neurons}, we divide the first quarter and segmented this period into three points for reward transition. We set the transition timings at \( t \in \{1N, 2N, 3N\} \), where \( N \) corresponds to roughly a third of the first quarter of the entire training duration. The specific value of \( N \), adjusted for each environment's episode length, is detailed in the Appendix A. We denote these phases as \( \mathscr{C}_1 \), \( \mathscr{C}_2 \), and \( \mathscr{C}_3 \), signifying the S2D or D2S reward transition points.

%Drawing inspiration from the foundational Sensorimotor Stage—often referred to as less than a quarter of the entire toddler learning period—where early development is characterized by toddlers predominantly engaging in free exploration~\cite{piaget1952origins, shonkoff2000neurons}, we segmented our training duration into three distinct parts. Using this first quarter as our baseline, transitions are delineated at \( t \in \{1N, 2N, 3N\} \). Specific settings for \( N \), tailored to each environment's episode length, can be found in Table 5 of the Appendix. We designate these phases as \( \mathscr{C}_1 \), \( \mathscr{C}_2 \), and \( \mathscr{C}_3 \), corresponding to the S2D or D2S transition junctures.

\begin{figure*}[t!]
     \centering
    \includegraphics[width=0.9\textwidth]{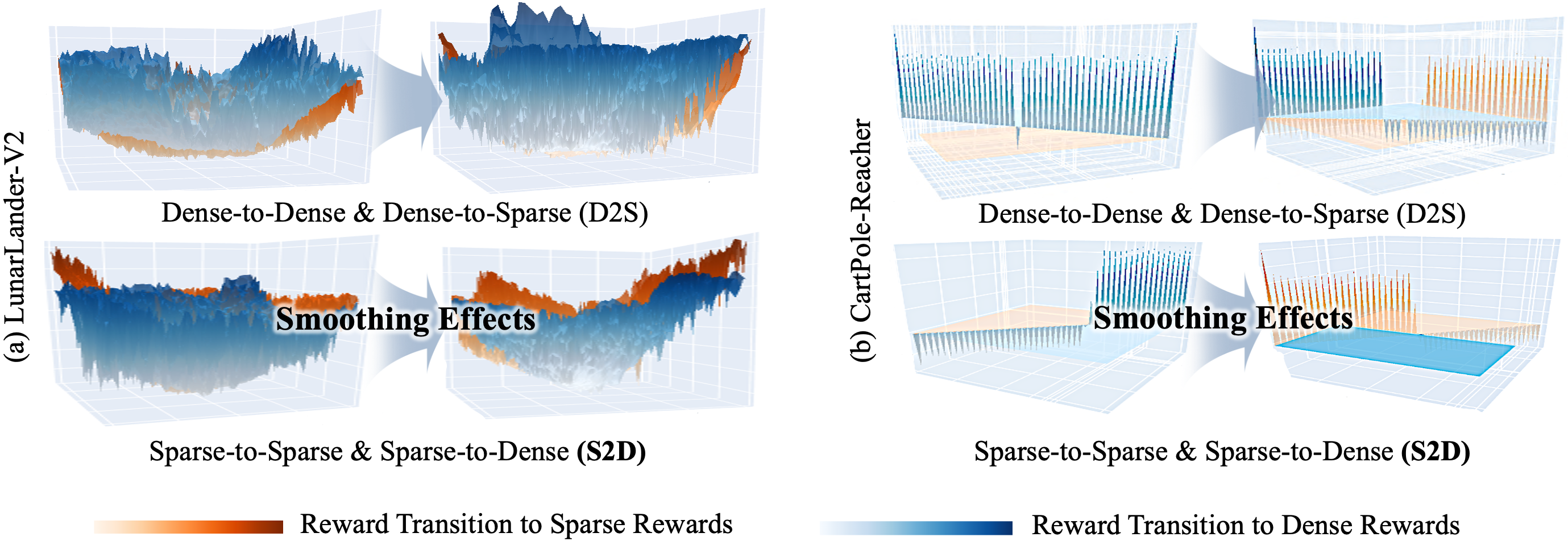}
    %\vskip -0.5em
    \caption{The 3D visualization illustrates the policy loss landscape following a reward transition, which begins with either a sparse or dense reward. It includes two sets of transitions: one transitioning from sparse-to-dense (S2D) and to sparse (Only Sparse), and the other from dense-to-dense (Only Dense) and to sparse (D2S). Notably, the S2D transitions often exhibited more distinct smoothing effects locally compared to others. These effects were noticeable after the transition at T=50 and T=2000 in LunarLander, and at T=3500 in Cartpole. Detailed 3D visualizations for environments are available in Appendix B.}
    %\vskip -0.5em
    \label{figure:visualpolicylossland} 
\end{figure*}

\begin{table*}[t]
\centering
\begin{adjustbox}{width=\textwidth}
\begin{tabular}{clccccccccc}
\toprule
\multirow{2}{*}{\thead{Task}} & \multirow{2}{*}{\thead{Reward\\Transition}} & \multicolumn{9}{c}{Number of updates after reward transition} \\
& & phase $\mathcal{P}_1$ & $\Delta\mathcal{P}_1$ & phase $\mathcal{P}_2$ & $\Delta\mathcal{P}_2$ & phase $\mathcal{P}_3$ & $\Delta\mathcal{P}_3$ & phase $\mathcal{P}_4$ & $\Delta\mathcal{P}_4$ & phase $\mathcal{P}_5$ \\
\midrule
\multirow{4}{*}{\thead{Lunar\\Lander}} & \underline{S2D}($\mathscr{C}_3$) & 0.031\stdv{0.01} & -0.004 & \bfseries 0.027\stdv{0.01} & +0.003 & \bfseries 0.030\stdv{0.01} & -0.008 & \bfseries 0.022\stdv{0.00} & -0.001 & \bfseries 0.021\stdv{0.00} \\
& D2S($\mathscr{C}_3$) & 0.043\stdv{0.02} & -0.008 & 0.035\stdv{0.00} & +0.004 & 0.039\stdv{0.01} & +0.004 & 0.043\stdv{0.02} & -0.009 & 0.034\stdv{0.01} \\
& only sparse & \bfseries 0.029\stdv{0.01} & 0.000 & 0.029\stdv{0.00} & +0.005 & 0.034\stdv{0.01} & -0.004 & 0.030\stdv{0.00} & +0.007 & 0.037\stdv{0.01} \\
& only dense & 0.039\stdv{0.01} & -0.007 & 0.032\stdv{0.01} & 0.000 & 0.032\stdv{0.01} & +0.005 & 0.037\stdv{0.01} & -0.002 & 0.035\stdv{0.01} \\
\midrule
\multirow{4}{*}{CartPole} & \underline{S2D}($\mathscr{C}_2$) & 0.028\stdv{0.00} & +0.003 & \bfseries 0.031\stdv{0.01} & 0.000 & \bfseries 0.031\stdv{0.00} & +0.005 & \bfseries 0.036\stdv{0.00} & -0.013 & \bfseries 0.023\stdv{0.00} \\
& D2S($\mathscr{C}_2$) & 0.033\stdv{0.01} & +0.014 & 0.047\stdv{0.04} & +0.001 & 0.048\stdv{0.01} & -0.002 &  0.046\stdv{0.02} & +0.001 & 0.047\stdv{0.02}\\
& only sparse & \bfseries 0.027\stdv{0.01} & +0.016 & 0.043\stdv{0.01} & -0.009 & 0.034\stdv{0.01} & +0.007 & 0.041\stdv{0.02} & -0.014 & 0.027\stdv{0.00} \\
& only dense & 0.033\stdv{0.01} & +0.021 & 0.054\stdv{0.02} & -0.013 & 0.041\stdv{0.02} & +0.004 & 0.045\stdv{0.02}& -0.001 &  0.044\stdv{0.01} \\
\midrule
\multirow{4}{*}{UR5} & \underline{S2D}($\mathscr{C}_3$) & 0.084\stdv{0.01} & +0.016 & 0.100\stdv{0.02} & -0.033 & 0.067\stdv{0.01} & +0.028 & 0.095\stdv{0.02} & -0.062 & \bfseries 0.033\stdv{0.01} \\
& D2S($\mathscr{C}_3$) & 0.060\stdv{0.03} & -0.007 & 0.053\stdv{0.02} & -0.005 &  0.048\stdv{0.02} & +0.011 &  0.059\stdv{0.02} & -0.003 &  0.056\stdv{0.01} \\
& only sparse & \bfseries 0.059\stdv{0.02} & -0.013 & \bfseries 0.046\stdv{0.00} & 0.000 & \bfseries 0.046\stdv{0.01} & +0.031 &  0.077\stdv{0.00} & -0.031 & 0.046\stdv{0.01} \\
& only dense & 0.079\stdv{0.02} & -0.012 & 0.067\stdv{0.02} & -0.009 &  0.058\stdv{0.02} & -0.005 & \bfseries0.053\stdv{0.01} & +0.005 & 0.058\stdv{0.01} \\
\bottomrule
\end{tabular}
\end{adjustbox}
\caption{Comparison of the depth of local minima in policy loss following reward transitions. A lower depth of local minima indicates a smoother terrain. Phase $\mathcal{P}_i$ indicates the number of updates, which are $(T =50,400,800,1200,1600)$ for LunarLander, and $(T = 50,1000,2000,3000,4000)$ for others.}
\label{table:reductiondepth}
\end{table*}

\subsection{Environment Details}

We assess the efficacy of reward dynamics across diverse conditions, encompassing state and visual observations as well as both discrete and continuous action domains, as seen in Appendix A. We evaluate the reward settings, including S2D reward transition, across a variety of goal-based tasks in well-established benchmark environments. As seen in Figure~\ref{allenvvv}, these include LunarLander~\cite{brockman2016openai}, CartPole, and UR5~\cite{todorov2012mujoco}. In Appendix A, we particularly detail challenging dynamics for UR5 and CartPole, featuring randomized placements of the agent, goal, and obstacles, termed the 'reacher' version. All agents have full access to the current state and are tested using the SAC~\cite{haarnoja2018soft} algorithm. We also adjusted the reward structure for both sparse and dense settings, with details in Appendix A.

% The UR5 and Cartpole are conducted in more challenging dynamic settings, which we have modified, termed as the ``-reacher version,'' shifting from fixed to randomized agent, goal, or obstacle placements, necessitating robust agent generalization as detailed in Appendix A.
 
\subsubsection{Environment for generalization.}
To measure the improvement in generalization, we design a challenging egocentric navigation task using the ViZDoom environment~\cite{kempka2016vizdoom}, depicted in Figure~\ref{allenvvv}. In the \textbf{Seen} environment (Appendix Figure 9-(a)), object locations are random, and walls have one of three textures. The \textbf{Unseen} environment (Appendix Figure 9-(b)) requires generalization to three new wall textures, distinct from the \textbf{Seen} environment. Here, We employed A3C~\cite{mnih2016asynchronous}.
 
\section{Results}

\subsection{Performance Enhancement}

\subsubsection{Sample efficiency and success rate.} 
Experiments are conducted in diverse environments with sparse rewards, and the results are shown in Figure~\ref{fig:main_exp},~\ref{fig:vizdoom} and Table~\ref{table:sharpness}. The agents in LunarLander, CartPole-Reacher, ViZDoom-Seen, and ViZDoom-Unseen environments achieve the lowest performance with default sparse rewards. S2D consistently outperforms all other baselines in these settings, achieving better sample efficiency.
% In UR5-Reacher, the task's sensitivity to reward density makes it notably harder with only sparse rewards than in other environments. Yet, the S2D strategy consistently outperforms other baselines in this challenging context.
Even in UR5-Reacher, which is notably more challenging with only sparse rewards than in other environments, the S2D strategy consistently outperforms other baselines. While algorithms based on intrinsic motivation often prioritized exploration over goal attainment in goal-oriented RL tasks, our approach exhibited exceptional results, exhibiting better exploration-exploitation tradeoff while being simple and universally applicable. Importantly, the outcomes associated with D2S consistently fall below those of S2D across all environments, underlining the efficacy of the S2D transition as a curriculum. 

%Moreover, in the grid world experiment, using PPO, S2D shows a success rate of $23.23 \pm 1.12$, Only Dense performs at $20.03 \pm 0.93$, Only Sparse reaches $18.9 \pm 0.96$, and D2S achieves $16.3 \pm 1.72$. We verify that S2D guidance outperforms all other baselines.

\subsubsection{Generalization performance.} 
The S2D reward transition outperforms other agents across all dynamic environments requiring generalization as seen in Figure~\ref{fig:vizdoom}-(a), (b). Particularly in ViZDoom-Unseen, where agents encounter drastic visual changes with the emergence of three previously unseen walls, the S2D transition achieves robust generalization and superior performance compared to other baselines.

\subsection{Post-Transition 3D Policy Loss Landscape} We note a marked smoothing effect at various update points after the reward transition, attributed to the reduction in the depth of local minima, as shown in Table~\ref{table:reductiondepth} and Figure~\ref{figure:visualpolicylossland}. This effect is predominantly observed under the S2D reward. Such smoothing could aid in overcoming local minima, possibly leading to wide minima. 3D policy loss landscapes of other reward baselines are visualized using the Cross-Density Visualizer in Appendix B. To further verify the smoothing quantitatively, we calculate the depth of local minima of $\pi_\theta$ function, which measures the average of differences between the  local maximum and minimum values for a number of updates after the reward transition. The results, as shown in Table~\ref{table:reductiondepth}, demonstrate that the depth primarily decreased for the S2D reward transition model.

\subsection{Results of Wide Minima} \label{wideminimaresult} We assess the end-of-training convergence of the neural networks guided by S2D using sharpness metrics and contrast this with the baselines. Areas of lower sharpness correspond to wide minima, which can enhance generalization performance. As demonstrated in Table~\ref{table:sharpness}, only the agents guided by S2D reward transition that converge to the widest minima exhibit superior performance in challenging environments.

\section{Discussion and Analyses}

In the following analyses, we clearly address the three pivotal questions raised in our experimental framework:

\subsubsection{Performance enhancement.} As shown in Figure~\ref{fig:main_exp}, Figure~\ref{fig:vizdoom}, and Table~\ref{table:sharpness}, S2D surpassed other reward baselines. In our experiments comparing intrinsic motivation methods, we observed that agents driven by such algorithms tend to prioritize state coverage over achieving specific goals in goal-oriented RL tasks. This suggests that while they may excel in exploration, the S2D transition approach more effectively balances exploration and exploitation, thus ensuring proper goal acquisition. Additionally, we explored the optimal timing for reward transition through ablation studies. While this timing is unique for each environment, it was, in all cases, near a quarter of the total training time. Particularly challenging tasks like UR5-Reacher required longer periods of free exploration compared to relatively simpler ones, such as LunarLander. This echoes the critical early learning phases observed in infants.

\subsubsection{Post-transition 3D policy loss landscape.} Our 3D visualizations reveal that S2D predominantly smooths the landscape. Although our main experiments are based on SAC, we also tested other algorithms such as PPO~\cite{Schulman2017ProximalPO} and DQN~\cite{mnih2013playing} for more comprehensive analyses. This smoothing effect was also uniquely seen with the S2D reward transition in additional gridworld experiments, as discussed in Appendix C. 

%Our quantitative analysis also highlighted the S2D transition's role in reducing the depth of local minima, enhancing the chances of parameters escaping from such traps. Crucially, a reduced depth of minima correlated with improved performance. For instance, we observed a distinctly gradual increase in UR5-Reacher as depicted in Figure~\ref{fig:main_exp}, which can be matched with the drastic depth reduction from 0.047 to 0.008 that occurred noticeably late, at phase $\mathcal{P}_5$ in Table~\ref{table:reductiondepth}.
%On the other hand, in CartPole-Reacher and LunarLander, performance surged shortly after the reward transition, corresponding to the S2D's low depth of local minima overall.

%Our analysis showed the S2D transition reduces local minima depth, aiding parameter escape, thus enhancing generalization. 
In UR5, minima depth for S2D decreased significantly from 0.095 to 0.033 in phase $\mathcal{P}_5$.
This may relate to its higher performance than only dense rewards particularly at a later stage, as in Figure~\ref{fig:main_exp}-(b). Conversely, in CartPole and LunarLander, performance quickly improved after the reward transition, reflecting S2D's overall low local minima.

\subsubsection{Correlation between wide minima and toddler-inspired reward transition.} In Table~\ref{table:sharpness}, S2D-guided agents converged to the widest minima with the highest performance in the LunarLander and CartPole-Reacher, where agents receiving only sparse rewards achieved nonzero performance.
Conversely, in the UR5-Reacher, agents with only sparse rewards showed zero performance. This implies that in sparse reward situations, premature convergence into wide minima can lead to gradient stagnation and retain low sharpness, with high variance. Yet, S2D, compared to only dense rewards, still posts the highest performance with the lowest sharpness, suggesting its alignment within wide minima.

% \vspace{-0.1in}
\section{Conclusion}
Inspired by toddler developmental learning, our research pioneers a shift from static, single-density to dynamic reward transitions in goal-oriented RL. This toddler-inspired approach demonstrates notable effects of transitions on learning dynamics in RL. We examine its implications across various scenarios, focusing on its efficiency. Using the Cross-Density Visualizer, we observe the primary smoothing effects on the policy loss landscape during the S2D transition. Sharpness metrics further confirm the smoothing effects of S2D, guiding agents towards wider minima to improve generalization. This blend of biological and artificial paradigms may lead to robust, high-performance learning systems.

% \vspace{-0.1in}
%Applying a toddler-inspired reward transition in RL, we observed phenomenological effects on the learning dynamics. Our research delves deeply into the implications of this transition across various scenarios, emphasizing its efficiency. Via Cross-Density Visualizer, we visualize the policy loss landscape during the reward transition and highlight its pronounced smoothing effects. Sharpness metrics further validate these distinct smoothing effects in S2D, steering the agent towards wide minima, thereby enhancing generalization. The examined approach, an amalgamation of biological and artificial paradigms, may instigate future works on learning systems that are both robust and performant.

%In our exploration, the Toddler-inspired S2D approach not only echoes natural learning processes but also aims to converge biological and artificial paradigms. Such an amalgamation promises RL systems that are not just algorithmically efficient but also inherently robust and adaptive, reshaping the very core of machine learning's future potential.

\textbf{Limitations.}
While our study focuses on understanding the implications of S2D reward transition, we haven't provided an automatic method for finding an optimal transition yet. Nonetheless, our preliminary research on criteria sets the stage for future development of automated methods.
%Moving forward, there are still challenges to be addressed regarding toddler-inspired reward transition.%and identifying an optimal reward transition.
% (위 문장 대체 시도... 아닌 것 같습니다) Our study focuses on understanding the implications of S2D reward transition how the timing of transition may relate to effective reward structure. % 음... 원석선배 말씀대로 시도해봤는데 잘 안 되네요
%While our study focuses on understanding the implications of S2D reward transition, we have not provided an automatic method for finding an optimal transition.
%Nonetheless, our preliminary research on potential reward transition criteria lays the foundation for developing more efficient automated detection methods in the future. 

\section{Acknowledgments}
The authors would like to express their sincere gratitude to Inwoo Hwang, Changhoon Jeong, Moonhoen Lee, and Dong-Sig Han for their insightful discussions and valuable suggestions on the early drafts of this paper. This work was partly supported by the IITP (2021-0-02068-AIHub/15\%, 2021-0-01343-GSAI/10\%, 2022-0-00951-LBA/15\%, 2022-0-00953-PICA/25\%) and NRF (RS-2023-00274280/10\%, 2021R1A2C1010970/25\%) grant funded by the Korean government.
%By leveraging the noticeable smoothing effects and examining sharpness variations during reward transitions, we may aim to enhance the detection and further improving the practicality and efficacy of our models. 

\bibliography{aaai24}

%\bibliography{reference_neurips}

%\iffalse
\appendix
% \documentclass[./Main.tex]{subfiles}

% \begin{document}
\onecolumn
\suppletitle{Appendix.\\Unveiling the Significance of Toddler-Inspired Reward Transition in Goal-Oriented Reinforcement Learning}
\setcounter{page}{1}

\begin{mdframed}[frametitlealignment=\centering,leftline=false,
  rightline=false]
\begin{itemize}[leftmargin=*]
\item \textbf{Section A:} This section provides further details about the experiments conducted in the main text and the appendices.

\item \textbf{Section B:} This section presents comprehensive results of visualizing the 3D policy loss landscape after the stage-transition for Toddler-inspired S2D Reward Transition and various baselines, which is related to the section: Visualizing Post-Transition 3D Policy Loss
Landscape: Cross-Density Visualizer in the main text.

\item \textbf{Section C:} This section features additional experiments and analyses, as well as further visualizations of the 3D policy loss landscape for various algorithms in the gridworld.
\end{itemize}
\end{mdframed}

\section{Section A: Experimental Details}
\subsection{A.1 Comparison of Overall Experimental Setup}
%\vspace{-0.5em}
\begin{table}[h]
    \renewcommand*{\arraystretch}{1.1}
    \begin{center}
        \caption{An overall comparison of experimental environments. Those above the double line relate to environments from the main text, while those below correspond to those in the appendices. Each environment and reward was modified to adapt to our experiments for Toddler-inspired S2D reward transition, with details provided in the respective environment setup descriptions.}
        % \label{table1}
        \label{table:detailenv}
        %\vspace{0.25em}
        \begin{adjustbox}{width=\textwidth}
            \begin{tabular}{c|cccc}
            \Xhline{3\arrayrulewidth}
            \multirow{2}{*}{Environment}& Task  & Difficulty Settings &Environments Type & Input\\
            & \# of Stages & Point of View & Action Space & Observation Types\\
            \Xhline{3\arrayrulewidth}
            \rowcolor[HTML]{EEEEEE}
            &LunarLander-V2&-&2D& Coordinate \& Velocity \& Angle \& Boolean flag value \\
            \rowcolor[HTML]{EEEEEE}
            \multirow{-2}{*}{\cellcolor[HTML]{EEEEEE} OpenAI Gym\citep{todorov2012mujoco}}&2-stage  & Allocentric View  &Continous& State-based RL\\
            \multirow{2}{*}{MuJoCo\citep{todorov2012mujoco}} &CartPole-Reacher&-&3D& Joint Value \& Goal Position\\
            &2-stage &Allocentric View&Continuous& State-based RL\\
            \rowcolor[HTML]{EEEEEE}
            &UR5-Reacher&-&3D& Joint Value \& Goal Position\\
            \rowcolor[HTML]{EEEEEE}
            \multirow{-2}{*}{\cellcolor[HTML]{EEEEEE} MuJoCo\citep{todorov2012mujoco}}&2-stage &Allocentric View&Continuous& State-based RL\\
            &  Seen \& Unseen & - &3D& RGB-D\\
            \multirow{-2}{*}{ViZDoom\citep{kempka2016vizdoom}} & 2-stage &Egocentric View&Discrete& Visual RL\\
            \hline
            \hline
            \rowcolor[HTML]{EEEEEE}
            &Shelf-delivery&Level3&2D& Internal state of the surrounding tiles\\
            \rowcolor[HTML]{EEEEEE}
            \multirow{-2}{*}{RWARE\citep{papoudakis2021benchmarking}}&Non-humanoid&Allocentric View& Discrete&3-stage \& State-based RL\\
            \rowcolor[HTML]{FFFFFF}
            &  Navigation & - &2D&  Position-based value\\
            \rowcolor[HTML]{FFFFFF}
             \multirow{-2}{*}{\cellcolor[HTML]{FFFFFF} Gridworld} &2-stage &Allocentric View&Discrete& State-based RL\\
            \Xhline{3\arrayrulewidth}
            \end{tabular}
        \end{adjustbox}
    \end{center}
    %\vspace{-1em}
\end{table}

\begin{table}[h]
%\vspace*{-6em}
    \centering
    \renewcommand*{\arraystretch}{1.1}
    \centering
    \caption{Detailed settings for hyperparameter $N$.
    %$N$ is each stage transition onset following the total number of frames of training for each environment.
    $N$ is the number of frames of training for the environment after which the stage transition occurs for the respective transition setting ($\mathscr{C}_1$, $\mathscr{C}_2$, and $\mathscr{C}_3$).
    In case of ViZDoom experiments, $N$ indicates the number of updates. For Gridworld-DQN,  $\mathscr{C}_1$=100, $\mathscr{C}_2$=200, and $\mathscr{C}_3$=300 episodes.
    }
     %\vskip -0.1in
    \begin{adjustbox}{width=0.7\textwidth}
        \begin{tabular}{c|cccc}
        \Xhline{3\arrayrulewidth}
        \multirow{1}{*}{Environment} & Total \# of Train & 1$N$(${\mathscr{C}_1}$) & 2$N$(${\mathscr{C}_2}$) &  3$N$(${\mathscr{C}_3}$)\\
        \Xhline{3\arrayrulewidth}
         \rowcolor[HTML]{EEEEEE} 
        LunarLander-V2  & 1M frames& 100k & 200k & 400k \\
   
        ViZDoom-Seen \& Unseen  & 1M frames& 50k & 100k & 250k \\
         \rowcolor[HTML]{EEEEEE} 
        CartPole-Reacher  & 12k episodes & 1k & 2k & 3k\\
      
        UR5-Reacher  & 25k episodes & 1k & 2k & 3k \\
        \hline
        \hline
        \rowcolor[HTML]{EEEEEE} 
        RWARE & 7M frames& 1M & 2M & 3M \\
        
        % ViZDoom-FourObj  & 7M frames& 1M & 2M & 3M \\
         \rowcolor[HTML]{FFFFFF} 
        Gridworld& 25k episodes& 3k & 5k& 7k \\
        \Xhline{3\arrayrulewidth}
        \end{tabular}
    \end{adjustbox}
   %\vskip -0.1in
   \label{table:hyper2}
\end{table}

\subsection{A.2 Reward Transition Hyperparameters}
\begin{figure}[thb]
\centering
    \includegraphics[width=1\columnwidth]{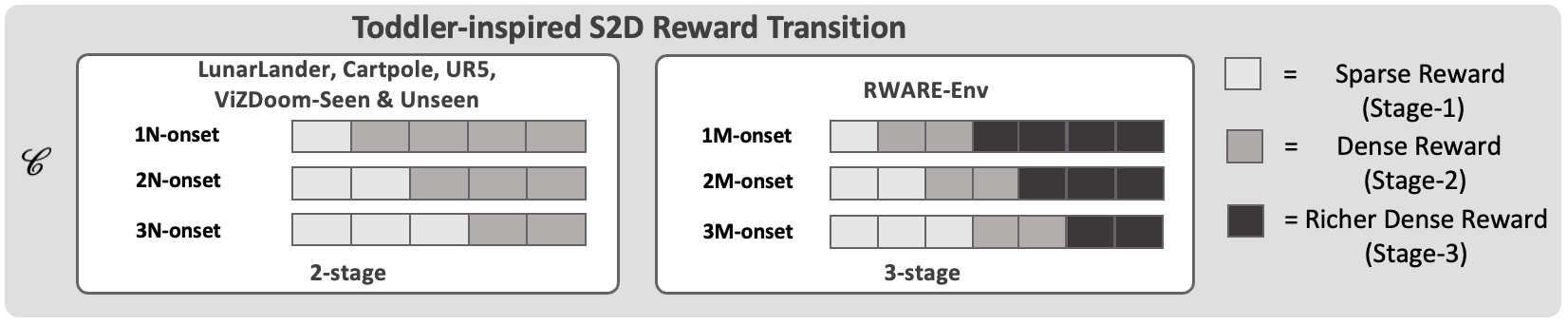}
   % \vskip -0.1in
    \caption{Visualization of the overall setup, including the number of stages and the transition times, in Toddler-inspired S2D experiments across all environments.}
   \label{figure:Toddler-inspired S2D}
\end{figure}

%\vspace*{-0.3em}
In Table~\ref{table:hyper2} and Figure~\ref{figure:Toddler-inspired S2D}, we have documented the specific point at which the agent transitioned to dense reward stages for each environment with a number of stages.

\subsection{A.3 Model Hyperparameters}
\label{app:hyper}
The Table~\ref{table:hyper3} provides information on the hyperparameters for each environment used in our study.

\begin{table}[h]
    \renewcommand*{\arraystretch}{1.1}
    \centering
    \caption{Hyperparameters for our various experiments for Figure~\ref{fig:main_exp} and Appendices. For visualizing the policy loss landscape in LunarLander, we employed discount factors $\gamma$ of both 1 and 0.99.}
    \begin{adjustbox}{width=\textwidth}
        \begin{tabular}{c|cccccccc}
        \Xhline{3\arrayrulewidth}
        Hyperparameters & LunarLander & CartPole-Reacher & UR5-Reacher & ViZDoom-S\&U & RWARE & Gridworld \\
        \Xhline{3\arrayrulewidth}
        \rowcolor[HTML]{EEEEEE} 
        RL algorithms &  SAC & SAC & SAC & A3C & PPO & PPO \\
        Learning Rate & 3e-4 & 0.0007 & 0.0007 & 7e-5 & 5e-4 & 5e-4\\ 
        \rowcolor[HTML]{EEEEEE} 
        Value Function Coefficient & 3e-4 & - & - & 0.5 & - & 5e-4 \\
        Discount Factor & 0.99 & 0.99 & 0.99 & 1.0 & 0.99 & 0.99 \\
        \rowcolor[HTML]{EEEEEE} 
        Batch Size & 128 & 128 & 128 & - & 10 & 128 \\
        Optimizer & Adam & Adam & Adam & Adam & Adam & Adam \\
        \rowcolor[HTML]{EEEEEE} 
        Maximum \# of Steps & 500 & 200 & 500 & 50 & 150 & 50 \\
        Entropy Coefficient & 0.2 & Auto & Auto & 0.1 & 0.01 & 0.03 \\
        
        \Xhline{3\arrayrulewidth}
        \end{tabular}
    \end{adjustbox}
    \label{table:hyper3}
\end{table}

%stage 3.

%\subsection{OpenAI Gym : Simple, pythonic, and capable of representing general RL problems,}

\subsection{A.4 Environment Details}
All reward-driven baselines and S2D reward were run with more than 4 random seeds for each environment. Our experiments were conducted using four NVIDIA GeForce RTX 3090 GPUs, or two NVIDIA GeForce RTX 2080ti GPUs, with the CPU being an AMD Ryzen Threadripper 3960X 24-Core Processor, and utilizing 188GB of RAM.

\subsubsection{Design of sparse to potential-based dense reward.}
\label{app:reward_setting_mujoco}
In a sparse-to-dense curriculum, an agent is eventually trained in a dense environment. Here, if a dense reward is not properly defined, the optimal agent in the dense environment may not be optimal in the base (sparse) environment. Therefore, an additional reward in the dense environment must be defined properly so that policy optimality is preserved.
In order for optimality to be preserved, we utilize potential-based reward shaping (PBRS) \cite{Ng1999PolicyIU, harutyunyan2015expressing}. In PBRS, an additional reward is given as follows:\footnote{While $\Phi$ is the notation commonly used for potential function, we have used $\psi$ throughout our paper.}
\begin{equation}
F_i(s, a) = \mathbb{E}_{s' \sim \mathcal{P}(s, a)}[\gamma \Phi_{i}(s') - \Phi_{i} (s)],
\end{equation}
where $\Phi_i : \mathcal{S} \rightarrow \mathbb{R}$ is 
 an arbitrary function called a \emph{potential function} at stage $i$.

By utilizing the additional reward function $F_i$, optimality of the policy is preserved, \emph{i.e.}, optimal policy $\pi^*$ with respect to reward function $\mathcal{R}_{i}+F_i$ is still optimal with respect to $\mathcal{R}_{i}$, as indicated below \cite{harutyunyan2015expressing}. We note that $R_i + F_i$ gives a denser reward than $R_i$.
\begin{align*}
\small
Q(s, a)
&=\mathbb{E}_{\mathcal{P}, \pi}\left[\sum_{t=0}^{\infty} \gamma^t\left(\mathcal{R}^t_{i}+F^t_{i}\right) \mid s_0=s\right]\\
&=\mathbb{E}_{\mathcal{P}, \pi}\left[\sum_{t=0}^{\infty} \gamma^t\left(\mathcal{R}^t_{i}+\gamma \Phi_{i}\left(s_{t+1}\right) - \Phi_{i} (s_t)\right) \mid s_0=s\right]
=\mathbb{E}_{\mathcal{P}, \pi}\left[\sum_{t=0}^{\infty} \gamma^t \mathcal{R}^t_{i} - \Phi_i (s_0)\right].
\end{align*}
% Since $R_i + F_i$ gives a denser reward than $R_i$, we can see that the guidance designed with PBRS is a \emph{PS2D-curriculum} by \cref{def:Anti_curriculum}.

In the sparse reward setting ($\mathcal{M}_1$), the agent only receives a reward when it successfully reaches the goal within a specific distance threshold. In the dense reward setting ($\mathcal{M}_2$, $\mathcal{M}_3$), the agent receives a potential-based dense reward in addition to the goal-based reward. The dense reward $\psi(\cdot)$, shown in Equation~\ref{eq:dense_guidance}, is determined by the agent's proximity to the goal, which is measured using a positive value based on the L2-distance between the agent's current position $s \in \mathcal{S}$ and the goal $g \in \mathcal{G}$:
\begin{equation}
    \psi(s) = \diam(\mathcal{S}) - ||s - g||_2,
    \label{eq:dense_guidance}
\end{equation}
where $\mathcal{S}$ is set of states, $\mathcal{G} \subseteq \mathcal{S}$ is set of goal states, and $\diam(\cdot)$ denotes the diameter of a set.

%---- Add the Explanation of Table 6 준석
From a potential-based reward shaping perspective, Table~\ref{tab:my-table} showcases the precise formulations for both sparse and dense rewards across various experimental environments, utilizing distance as the pivotal potential function in the reward design.

\begin{table}[H]
   \caption{
   Reward setting formulation in the environments used for the experiments.
   Each cell indicates the condition at which a respective sparse or dense reward is provided to the agent.
   The specific reward values are outlined in the text for each task description.
   }
   \centering  % This line centers your table
   %\vspace{0.25em}
   \begin{adjustbox}{width=\textwidth, totalheight=\textheight, keepaspectratio}
      \begin{tabular}{|c|c|c|c|c|c|}
         \hline
         \rowcolor[HTML]{EEEEEE} 
         {\color[HTML]{000000} \textbf{Reward}} & {\color[HTML]{000000} \textbf{LunarLander-V2}} & {\color[HTML]{000000} \textbf{CartPole-Reacher}} & {\color[HTML]{000000} \textbf{UR5}} & {\color[HTML]{000000} \textbf{ViZDoom-Seen}}& {\color[HTML]{000000} \textbf{ViZDoom-Unseen}} \\ \hline
         \textbf{Sparse} &  $||s-g||_2 < 1$ &  $||s-g||_2 < 0.02$ & $||s-g||_2 < 0.02$ &$||s-g||_2 < 0.0075$ & $||s-g||_2 < 0.0075$\\ \hline  % Added empty cells
         \textbf{Dense} & \multicolumn{1}{c|}{ $\gamma \psi(s_{t+1}) - \psi(s_t)< 0.3$ } &\multicolumn{2}{c|}{ $\gamma \psi(s_{t+1}) - \psi(s_t)<1$ } &\multicolumn{2}{c|}{ $\gamma \psi(s_{t+1}) - \psi(s_t)<0.14$ } \\ \hline  % Changed multicolumn to span 5 columns
      \end{tabular}%
   \end{adjustbox}
   \label{tab:my-table}
\end{table}

\subsubsection{OpenAI Gym: LunarLander-V2.}
The lander begins each episode in mid-air, equipped with a random initial velocity and orientation. The primary objective for the agent is to skillfully maneuver the lander's engines, guiding it to touch down between the two flags. Ideally, this landing should be at the center of the pad, as vertically aligned as possible, and with minimal speed to ensure a gentle touchdown. Rewards are granted for skillful approaches: transitioning from the screen's top to the landing pad, achieving a gentle landing with minimal speed, and for each leg that makes ground contact. Conversely, penalties apply for excessive main engine use, encouraging fuel conservation, and significant penalties are incurred for crashes or landing far from the intended pad. For our experiments, we employed the environment's default rewards as the sparse reward structure and incorporated distance-based, potential-driven dense rewards as part of our Toddler-inspired S2D reward transition, detailed in the above design of reward transition. 

Moreover, the potential-based reward shaping we introduced is rooted in a distance-based potential function. Unlike other goal-conditioned environments, in the LunarLander scenario, there is a possibility of receiving rewards for landing at any point on the ground. As such, when the lander approaches within a distance of 0.3 from the ground (where 1.0 indicates the entire screen height), as seen in Table~\ref{tab:my-table}, an additional potential-based reward per frame is granted.
 
%Luna Lander was trained with the SAC algorithm for 1M frames, and the random action period was set to the first 200 episodes, with a max step of 500, a batch size of 128, and a replay buffer size of 100000 for each episode. 

%\subsubsection{Ant-reacher Task}
%\emph{Ant} variants are environments where an 8-DoF Ant agent with four legs manipulates its joints to perform a variety of tasks. In Ant Reacher, the ant agent moves in a box-shaped space, and is given a goal to reach. The goal is randomly generated in the box every episode, and the agent starts from a random position distant from the goal. The default reward setting of the environment is extremely sparse \emph{i.e.} reward is given only when the agent reaches the goal.

\subsubsection{MuJoCo: CartPole-Reacher \& UR5-Reacher tasks.}
\label{mmp}

\begin{figure}[h!]
    \begin{center}
    \includegraphics[width=0.8\textwidth]{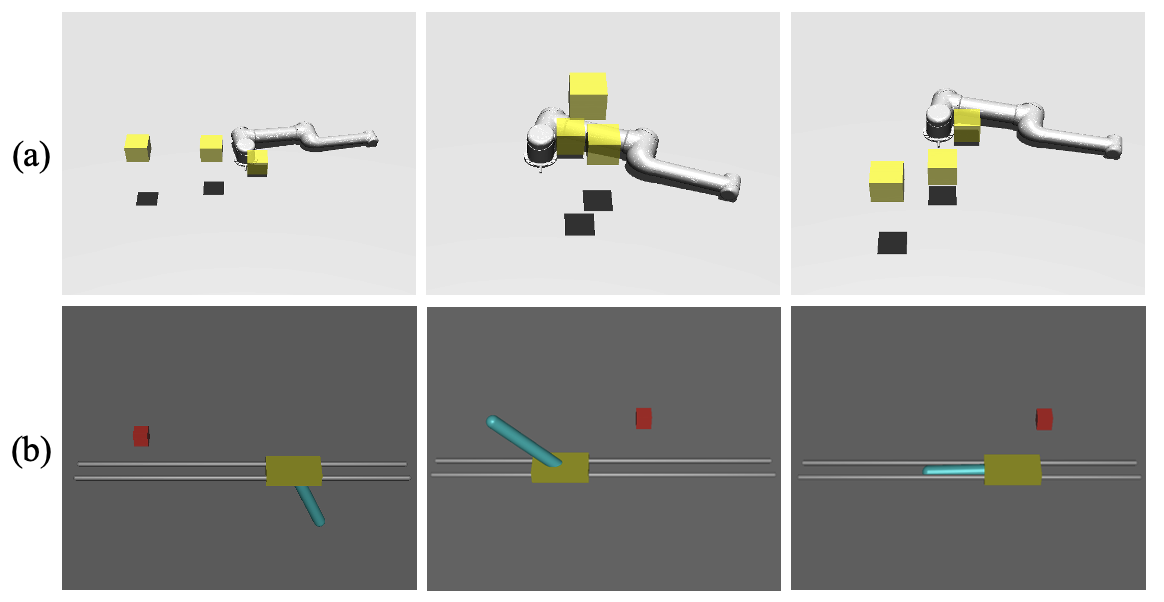}
    \caption{Examples of environments where goals randomly spawn. (a) UR5-Reacher. (b) CartPole-Reacher. }
    \label{ur5cart}
    \end{center}
\end{figure}

\begin{figure}[!htb]
    \begin{center}
    \includegraphics[width=0.8\textwidth]{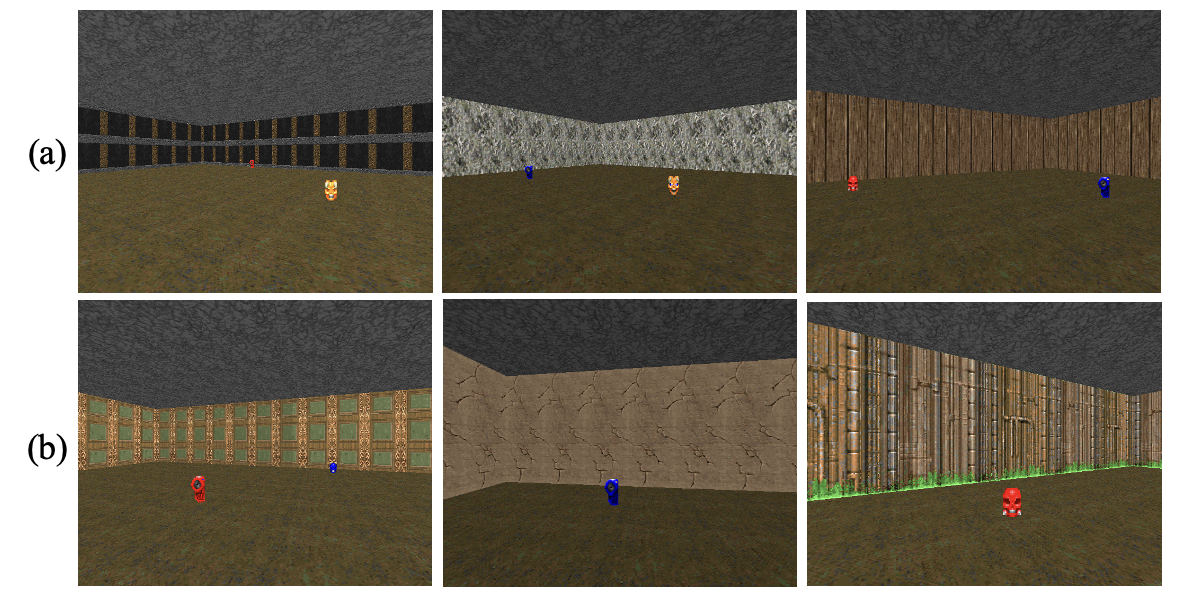}
    \caption{Egocentric views of a ViZDoom agent in environments with various walls and objects. (a) Three walls in ViZDoom-Seen. (b) Three walls in ViZDoom-Unseen.}
    \label{fig:viz-seen}
    \end{center}
\end{figure}

To evaluate and compare the effectiveness of Toddler-inspired S2D reward transition in a variety of tasks, we used MuJoCo \cite{todorov2012mujoco} engine, a widely used physics engine to simulate manipulation in virtual physical environments. Experiments were conducted in challenging continuous control tasks where a goal is randomly given each episode, and the agent must reach the goal. 

CartPole-Reacher is a task where an agent moves the cart along the horizontal line to make the connected pole stand for as long time as possible. Since the agent easily reaches the highest performance in the default task, we made it more challenging by setting a goal that the end of the pole must be within a radius less than a specific threshold. The goal is sampled in the upper side of the horizontal line where it is possible to reach the pole as seen in Figure~\ref{ur5cart}-(b).

UR5-Reacher is a six-degree-of-freedom arm with six joints that each allow movement along one degree of freedom. This flexibility enables the arm to reach a large number of positions and orientations but also makes controlling the arm a challenging task. The UR5-Reacher is often simulated as an environment where an agent must learn to control the arm. In our task, the agent needs to learn how to move the arm to a specific position. Our designed goal is randomly spawned for every episode, as seen in Figure~\ref{ur5cart}-(a).

For both UR5-Reacher and CartPole-Reacher, from the very beginning, potential-based dense reward is given based on the distance between the current state of the agent and the goal state as seen in Table~\ref{tab:my-table}.
Also, a constant time penalty is provided to encourage the agents to solve the tasks quickly.

\subsubsection{ViZDoom-Seen \& Unseen.}
% <introduction of vizdoom and the env by kim et al>
ViZDoom \cite{kempka2016vizdoom} is a simulator, based on the first-person shooter game Doom, that was developed to accelerate the reinforcement learning research.
In particular, we adopt and modify the egocentric navigation task designed by \cite{kim2021goal}, where the agent is placed in one corner of the square room at the start of the episode and must navigate to the correct object out of the two objects spawned in the room.
The correct ``goal'' object is specified every episode uniformly at random, which is known to the agent.
The two objects spawned are \textit{Card} and \textit{Skull}, each of them appearing as one of three color variants (Red, Blue, Yellow).
This prevents the agent from merely memorizing the objects' appearances according to their colors.
In addition, three wall textures are used in each of Seen and Unseen versions, and these two sets of wall textures are mutually exclusive.
% <egocentric navigation task>

% <observation: RGBD of 4 frames>
The agent receives a concatenation of 4 frames of RGB-D vision input.
Each vision channel has width and height of 42 and 42. The agent can choose among 3 discrete actions: turning clockwise, turning counterclockwise, and moving forward.
A single chosen action is repeated for 4 in-game frames.
Our manuscript refers to a single ``step'' within this environment as these 4 consecutive steps.
% <reward function>

As a sparse reward function, the agent receives a reward of 10 for arriving at the goal object, and a reward of -1 for arriving at the incorrect object.
Reaching either of these objects terminates the episode, which otherwise ends after 50 steps and grants a negative reward of -0.1.
For accelerating the training, a negative reward of -0.01 is provided every time step. To render this task visually complex, as well as to evaluate generalization of the trained agent, we use \textbf{Seen} and \textbf{Unseen} versions of the map as shown in Figure~\ref{fig:viz-seen} that differ in the set of wall textures used.
Dense reward is designed as an additional potential reward of 5.0 provided when the agent arrives within distance of 100 from the goal object.
Note that the map size is 700 by 700.

As the baseline, we use the A3C algorithm \cite{mnih2016asynchronous} and the architecture implemented by \cite{kim2021goal}\footnote{https://github.com/kibeomKim/GACE-GDAN} and \cite{kim2023sa,kim2023visual}.
We also note that all ViZDoom experiments were conducted on the two hardware settings.
Experiments on unified hardware settings will be further provided.
% <architecture: cite Kim et al, gace-gdan repo>
% <note that not all experiments were run on the same hardware settings>

%\subsubsection{\textcolor{red}{Design of reward transition on ViZDoom}} 

%\begin{table}[h!]

 %\vspace{-1em}
% \subsection{Limitation of Guidance}
%In our approach, it remains challenging to design appropriate potential functions tailored to each environment. This often necessitates the definition of hand-crafted features, which, if improperly designed, could potentially lead to entrapment in local optima or exert a negative impact. However, these are general issues in all reward shaping methodologies involving human bias. In an attempt to bridge this gap, we concentrate on potential-based reward shaping (PBRS) strategies, which is a field under active research to overcome these limitations. Specifically, we utilize these strategies in our novel \textit{Sparse reward-first, PBRS-later} (Toddler-inspired S2D) multi-stage framework. We aim to overcome these challenges more effectively than when solely using the Potential-based dense reward.

\newpage
\section{Section B: 3D Visualization of the Policy Loss Landscape After Stage
Transition}
\label{3dpolicydetal}
This study visualizes the 3D policy loss landscape after reward transitions from initially sparse or dense reward settings to two versions: one where sparse reward is provided and another where dense reward is provided.
This visualization is done within the same parameter space using \textit{Cross-Density Visualizer}.
As hyperparameters for Cross-Density Visualizer, the values for $\alpha$ and $\beta$ in $\tilde{\theta} = \theta + \alpha \mathbf{x} + \beta \mathbf{y}$ range from -10 to 10.
Consequently, Sparse-to-Dense (S2D) and Sparse-to-Sparse (Only Sparse) constitute one set, while Dense-to-Sparse (D2S) and Dense-to-Dense (Only Dense) make up another set. We observed a more numerous and distinct smoothing effect, especially under the reward transition of Toddler-inspired S2D. This smoothing effect could facilitate the surmounting of local minima, potentially leading to wider minima.
\subsubsection{Results.} In the Toddler-inspired S2D reward transition, a greater number of noticeable smoothing effects, such as the reduction in the depth of local minima, are observed. This is evident in the blue landscapes of Figures \ref{fig:lunar_S2D}, \ref{fig:cart_S2D}, and \ref{fig:UR5_S2D}, particularly in the sections below showing sparse-to-sparse and sparse-to-dense visualizations, in contrast to the D2S, Only Dense, and Only Sparse methods.

Based on our observation of reduced local minima depth, we hypothesize that this facilitates escape from local minima, improving generalization performance on wider minima. To test this, we measure the end-of-training convergence of Toddler-inspired S2D's neural network to wider minima using sharpness metrics and compare it with those of baselines. As shown in Table~\ref{table:sharpness}, Toddler-inspired S2D-guided agents converging to wider minima perform better in dynamic environments.
% This supports our conjecture that Toddler-inspired S2D guidance improves the likelihood of escaping local minima by smoothing the local loss landscape.
This suggests a correlation between the smoothing effect of the local loss landscape and the likelihood of escaping local minima.

\subsection{B.1 Additional Insights: Visualizing Policy Loss Landscape After Reward Transition}
To gain a deeper understanding of how reward transitions influence agent behavior, we visualized the policy loss landscape after transitions. This examination provides a nuanced view into the nature of the model's optimization landscape, revealing distinct challenges or advantages that can shape continuous learning.

\subsubsection{Distinguishing features of LunarLander-V2's landscape.}
What sets LunarLander-V2 as seen in Figure ~\ref{fig:lunar_S2D} apart from other environments is its unique reward distribution. In this environment, actions such as moving from the screen's top to the landing pad or achieving a resting state provide rewards ranging between 100 and 140 points. Adversely, actions like deviating from the landing pad or causing a crash lead to deductions. 
%This plethora of reward avenues renders agent's policy loss landscape smoother and less peak-ridden compared to others, a likely outcome of its diverse reward scenarios beyond primary goals.
We speculate that such a plethora of reward avenues is what renders the agent's policy loss landscape rounder and more gently sloped, and less peak-ridden compared to the landscapes in other environments.

\subsubsection{Pronounced peaks in CartPole-Reacher and UR5-Reacher.}
%In contrast, the CartPole-Reacher for Figure~\ref{fig:cart_S2D} and \ref{fig:cart_D2S} and UR5-Reacher-Reacher-Reacher for Figure~\ref{fig:UR5-Reacher_S2D} and \ref{fig:UR5-Reacher_D2S}, environments exhibit a more singular reward paradigm.
In contrast, the CartPole-Reacher and UR5-Reacher environments exhibit a more singular reward paradigm.
Rewards are focused and localized, leading to their policy loss landscapes showcasing distinct peaks, as shown in Figure~\ref{fig:cart_S2D} and \ref{fig:UR5_S2D} for CartPole-Reacher, and UR5-Reacher.
This is indicative of the few specific scenarios where agents earn positive feedback.

Through these visualizations, we not only discern the distinct reward mechanisms inherent to different environments but also appreciate how such structures sculpt the policy loss landscapes, influencing agent learning trajectories post-transition.

\iffalse% \newpage
\subsection{LunarLander-V2: Dense-to-Sparse (D2S) \& Dense-to-Dense (Only Dense)}
%\vspace{25pt}

\begin{figure}[H]
\begin{center}
%\vspace*{\fill}
\includegraphics[width=\textwidth, height=0.75\textheight, keepaspectratio]{LaTeX/icml2023/picture/newLunar-D2S.png}
\caption{Visualization of the 3D policy loss landscape after stage-transitions in dense-to-sparse (D2S, $\mathscr{C}_1$, red) and dense-to-dense (Only Dense, blue).
Each column shows the changes in policy loss landscapes at different time points in a single trial, where $T$ represents the time since transition.
% The policy loss is computed by averaging policy losses over a batch of transitions from the replay buffer, with each policy loss determined according to the parameter $\tilde{\theta}$. The formula $\tilde{\theta} = \theta + \alpha u + \beta v$ is used as proposed in \cite{li2018visualizing}, where $\alpha$ and $\beta$ are coordinates in the x and y axes and are normalized. The x and y axes are random orthogonal gradient directions, while the normalized Z axis indicates policy loss.
}
\label{fig:lunar_D2S}
%\vspace*{\fill}
\end{center}
\end{figure}
\fi

\newpage
\subsection{B.2 Detailed 3D Visualizations for All Baselines}
\subsection{LunarLander-V2: Dense-to-Sparse (D2S) \& Dense-to-Dense / Sparse-to-Dense (\textcolor{blue}{S2D}) \& Sparse-to-Sparse}
\begin{figure}[H]
\begin{center}
%\vspace*{\fill}
\includegraphics[width= 5\textwidth, height=0.75\textheight, keepaspectratio]{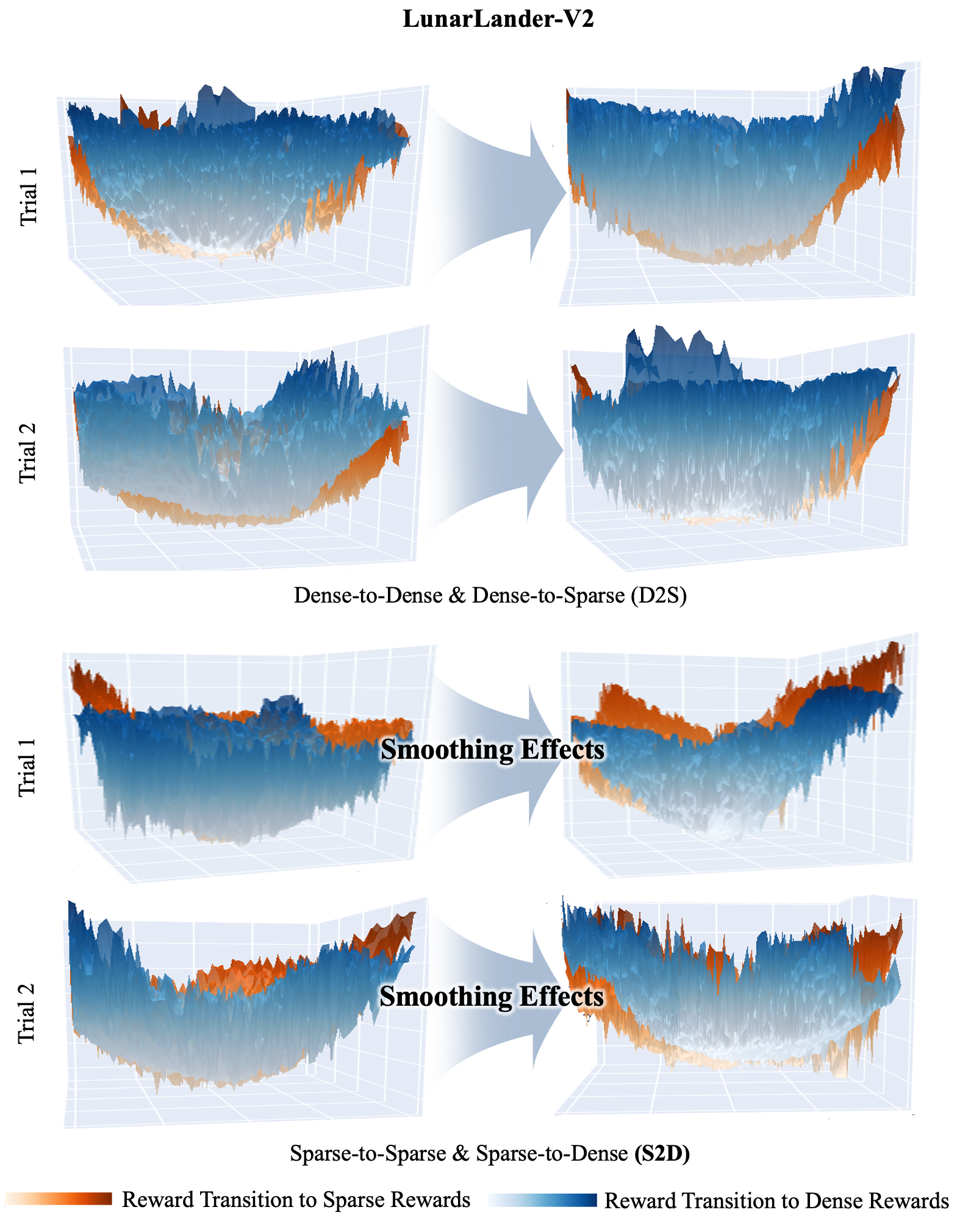}
\caption{ Visualization of the 3D policy loss landscape after reward transitions: The 'low' on the left side illustrates the landscape immediately following the transition, while the 'low' on the right side depicts the landscape around t = 2000 after the transition. The first two lines represent dense-to-sparse (D2S, $\mathscr{C}_1$, red) and dense-to-dense (Only Dense, blue) transitions. In contrast, the second set of lines represent sparse-to-dense (\textcolor{blue}{Toddler-inspired S2D}, $\mathscr{C}_2$, blue) and sparse-to-sparse (Only Sparse, red) transitions. Notably, the depth of local minima reduction (smoothing effects) is more pronounced under the Toddler-inspired S2D, observed across multiple update points after the reward transition.}
\label{fig:lunar_S2D}
%\vspace*{\fill}
\end{center}
\end{figure}

\newpage
\subsection{CartPole-Reacher: Dense-to-Sparse (D2S) \& Dense-to-Dense / Sparse-to-Dense (\textcolor{blue}{S2D}) \& Sparse-to-Sparse }
%\vspace{35pt}

\begin{figure}[H]
\begin{center}
%\vspace*{\fill}
\includegraphics[width=\textwidth, height=0.8\textheight, keepaspectratio]{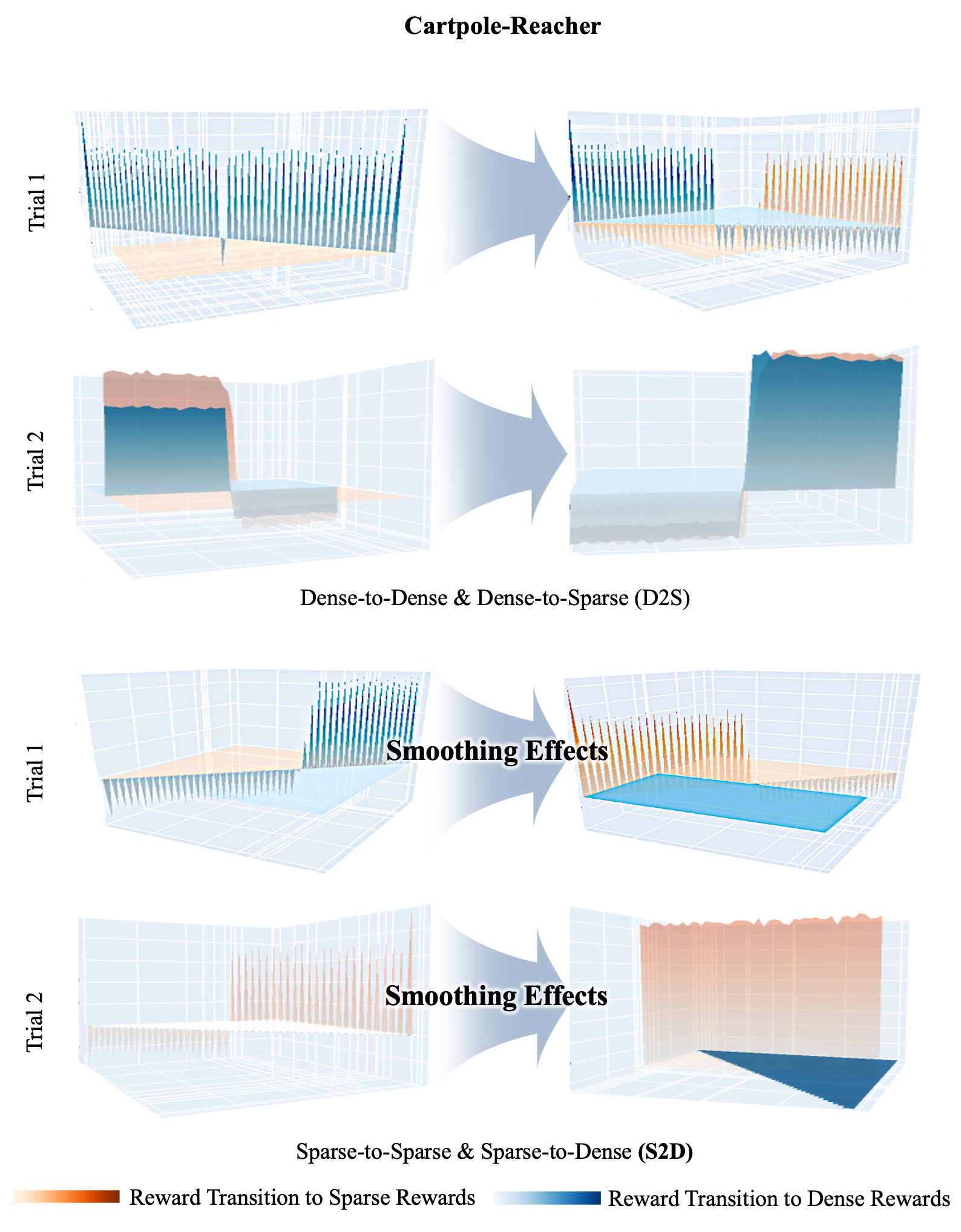}
\caption{In the visualization of the 3D policy loss landscape after reward transitions, significant smoothing effects were more frequently observed when applying the Toddler-inspired S2D (\textcolor{blue}{sparse-to-dense}, $\mathscr{C}_2$, blue) reward transition.}
\label{fig:cart_S2D}
%\vspace*{\fill}
\end{center}
\end{figure}

\newpage
\iffalse
\subsection{CartPole-Reacher: Dense-to-Sparse (D2S) \& Dense-to-Dense (Only Dense)}
%\vspace{35pt}
\begin{figure}[H]
\begin{center}
%\vspace*{\fill}
\includegraphics[width=\textwidth, height=0.8\textheight, keepaspectratio]{LaTeX/icml2023/picture/newCartpole-D2S.png}
\caption{Visualization of the 3D policy loss landscape after stage-transitions in dense-to-sparse (D2S, $\mathscr{C}_1$, red) and dense-to-dense (Only Dense, blue). It is observed that the depth of local minima fluctuates significantly with each update.}
\label{fig:cart_D2S}
%\vspace*{\fill}
\end{center}
\end{figure}
\fi

\newpage
\subsection{UR5-Reacher: : Dense-to-Sparse (D2S) \& Dense-to-Dense / Sparse-to-Dense (\textcolor{blue}{S2D}) \& Sparse-to-Sparse}
%\vspace{35pt}
\begin{figure}[H]
\begin{center}
%\vspace*{\fill}
\includegraphics[width=\textwidth, height=0.8\textheight, keepaspectratio]{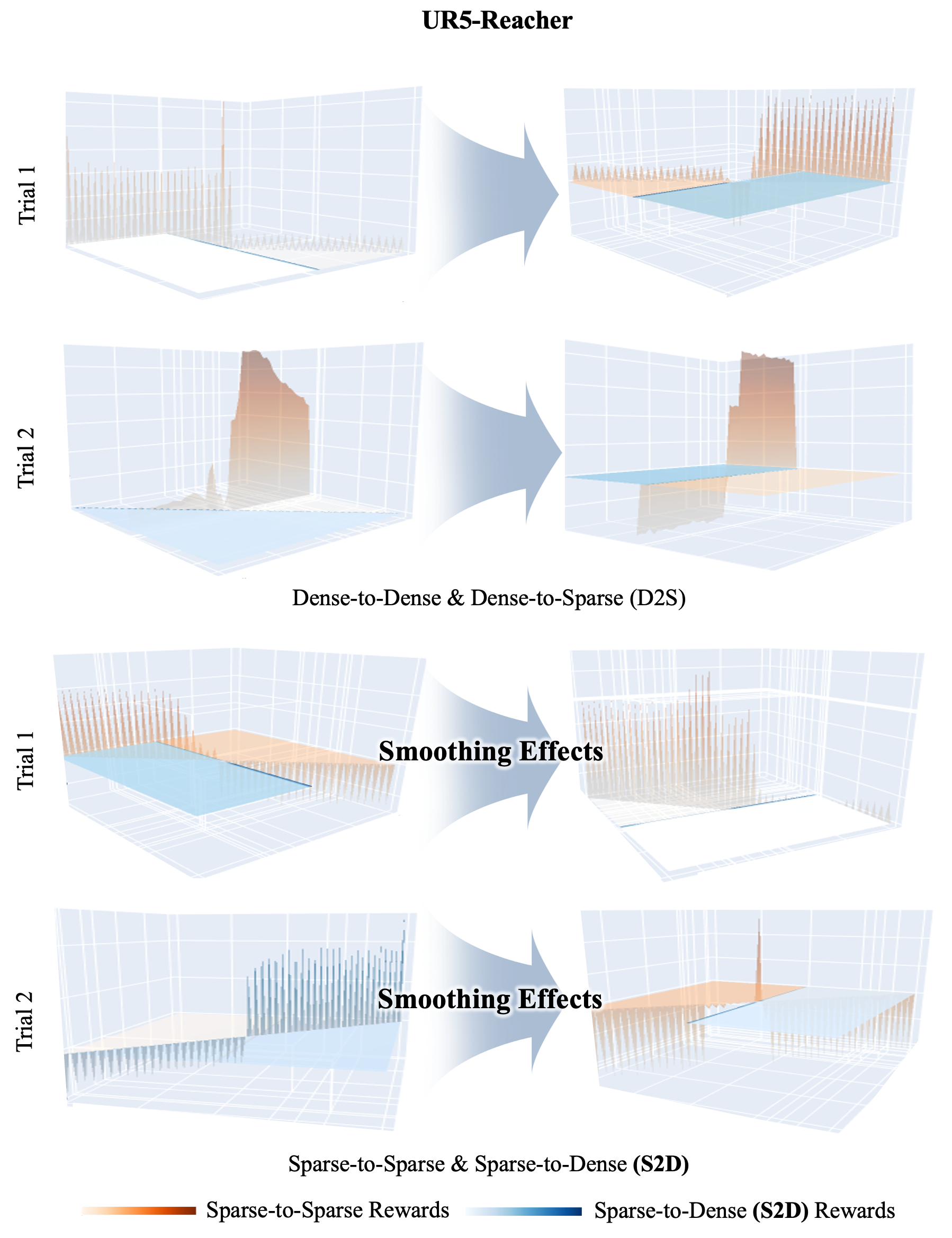}
\caption{Visualization of the 3D policy loss landscape after stage-transitions in sparse-to-dense (\textcolor{blue}{Toddler-inspired S2D}, $\mathscr{C}_3$, blue) and sparse-to-sparse (Only Sparse, red). Here, in the UR5-reacher, the improvement in performance of S2D over only dense becomes apparent later, and until that point, the performance of the two is generally not significantly different. Perhaps because of that, the Only Dense model also often showed a lower depth of local minima depending on the seed immediately after the reward transition. Yet, we observed even more smoothing effects due to the decrease in the depth of local minima under Toddler-inspired S2D reward transition, even compared to other baselines.}
\label{fig:UR5_S2D}
%\vspace*{\fill}
\end{center}
\end{figure}

\newpage
\iffalse
\subsection{UR5-Reacher: Dense-to-Sparse (D2S) \& Dense-to-Dense (Only Dense)}
%\vspace{35pt}
\begin{figure}[H]
\begin{center}
%\vspace*{\fill}
\includegraphics[width=\textwidth, height=0.8\textheight, keepaspectratio]{LaTeX/icml2023/picture/newUR5-D2S.png}
\caption{Visualization of the 3D policy loss landscape after stage-transitions in dense-to-sparse (D2S, $\mathscr{C}_3$, red) and dense-to-dense (Only Dense, blue). Similar to what is seen in CartPole-D2S and only dense scenario, we cannot observe stable smoothing effects consistent with the number of updates.}
\label{fig:UR5_D2S}
%\vspace*{\fill}
\end{center}
\end{figure}
\fi

\newpage
\section{Section C: Additional Experiments of Toddler-Inspired S2D Reward Transition in Various Environments}
\label{app:exp}

\subsection{C.1 Shelf Delivery Tasks in RWARE: Experiments on a Suboptimal Reward Structure with Three-Stage Guidance}

\subsubsection{Task settings.}
\begin{figure*}[t!]
    \centering
    \includegraphics[width= 0.9 \textwidth]{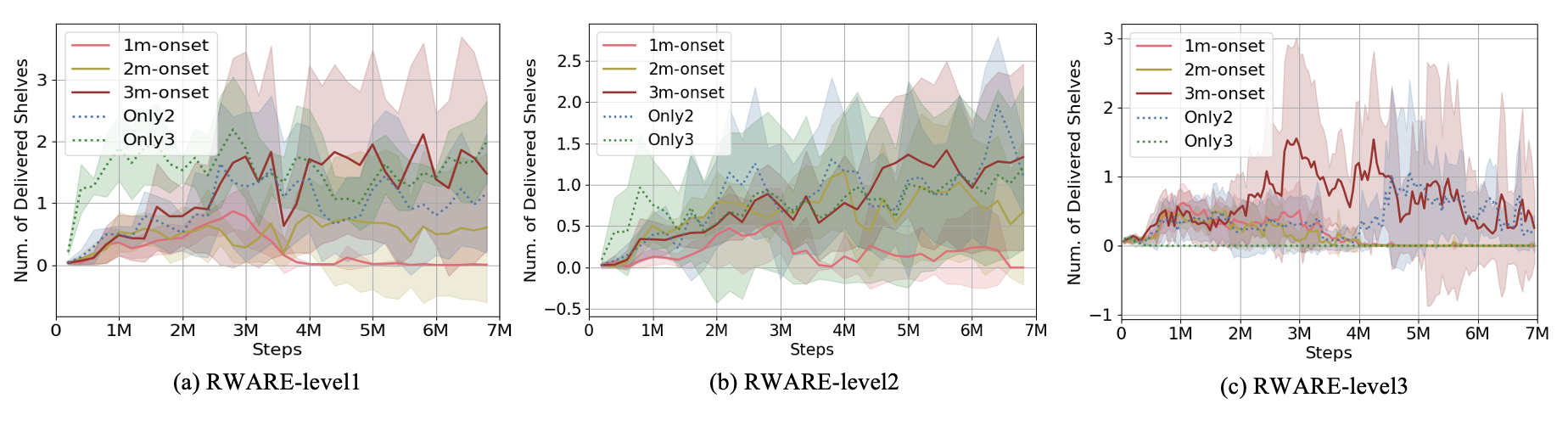}
    \caption{Performance of the agents in RWARE tasks.
    Vertical axis indicates the average number of requested shelves successfully delivered to the goal per episode.
    Horizontal axis indicates the number of time steps.
    The solid lines show the performances of Toddler-inspired S2D reward transition agents $(\mathscr{C}_1)$, $(\mathscr{C}_2)$, and $(\mathscr{C}_3)$, while the dotted lines show the performances of only dense reward agents $(\mathscr{C}_4)$.
    The mean and standard deviation of each agent in each level across five trials are shown as lines and shaded regions respectively.
    $(\mathscr{C}_3)$ agent shows one of highest-performing learning curves in the three environments, even when the Stage-3 guidance may be unhelpful as in Level 3.
    % The Toddler-inspired S2D guidance model, as well as the only dense guidance model, has a substantial standard deviation, so this is not a phenomenon because the Toddler-inspired S2D method is unstable.
    Both the Toddler-inspired S2D reward transition agents and the only dense reward agents show substantial standard deviation, which we claim is due to the difficulty of these RWARE tasks, which involve subgoals.
    }
   % \vskip -1em
    \label{fig:resultG2}
\end{figure*}

As seen in Figure~\ref{fig:resultG2} and ~\ref{Rwaretrajectory}, We conduct additional experiments on grid-world RWARE environments with discrete state-action space. Modified from RWARE \cite{papoudakis2021benchmarking} to be a single-agent setting, each environment contains a mobile agent and several rows of shelves (\textcolor{blue}{blue}), some of which are randomly sampled and ``requested'' (\textcolor{red}{red}) to be delivered.
The agent has an available action space of \{\textit{MoveForward, TurnLeft, TurnRight, Load/Unload, Noop}\}.
The agent observes only the information of tiles within 3-by-3 square area centered at the agent.
The agent must carry and deliver requested shelves to the goal location (\textcolor[HTML]{3d8c40}{green}).
This can be deemed as a task with subgoals, where the agent must first reach the tile with the requested shelf, pick up the shelf and carry it to the goal, and then return the shelf back to its original position.

We designate separate environmental settings for three difficulty settings: Level1 contains 8 shelves, 3 of which are requested; Level2 contains 10 shelves, 3 of which are requested, leading to sparser reward; Level3 is a vast environment with a total of 32 shelves, 16 of which are requested.
The space to explore is magnified as the difficulty increases.
These environments are illustrated in Figure~\ref{Rwaretrajectory}.

\subsubsection{Design of guidance.} Reward functions for three different stages of guidance are naturally designed according to this sequence of subtasks: Stage-1 guidance provides +1.0 reward when the agent successfully carries a requested shelf to a goal location; Stage-2 guidance provides reward for the delivery and returning the shelf to its original location; Stage-3 guidance provides an additional reward for picking up a requested shelf.

\subsubsection{Results on RWARE environments.}
The purpose of this study is to examine the effect of different Toddler-inspired S2D reward transition on learning the tasks in RWARE that involve a sequence of \textit{subgoals} in order to clearly observe the sub-optimal situation following the Toddler-inspired S2D models near the transition times.
The experiments investigated the Toddler-inspired S2D reward transition models at the timing of three different stage transitions ( $\mathscr{C}_1$, $\mathscr{C}_2$, and $\mathscr{C}_3$), as well as only dense reward applied to learn with only Stage-2 ($\mathscr{C}_4$) or Stage-3 ($\mathscr{C}_5$) guidance provided.
The transition times are the same as those specified in Figure~\ref{figure:Toddler-inspired S2D}, with $N$ being set to 1 million (1M) time steps.
PPO \cite{Schulman2017ProximalPO} was used as the RL algorithm, and the means and standard deviations were recorded across five trials.
We analyze the results below:

\begin{itemize}
\item \textbf{Level1.} (Figure \ref{fig:resultG2}-(a)).
Among the Toddler-inspired S2D reward transition agents, the $\mathscr{C}_3$ agent performs the best, while the $\mathscr{C}_1$ agent fails to deliver any shelves after the 4M time step.
While the Only 2 ($\mathscr{C}_4$) and Only 3 ($\mathscr{C}_5$) agents achieve remarkable performances, $\mathscr{C}_3$ agent performs on par with, if not better than, these only dense reward agents.

\item \textbf{Level2.} (Figure \ref{fig:resultG2}-(b)).
Similarly to the results in Level-1, the $\mathscr{C}_3$ agent performs the best, followed by $\mathscr{C}_2$ agent.
The Only 2 ($\mathscr{C}_4$) and Only 3 ($\mathscr{C}_5$) agents achieve performances similar to that of $\mathscr{C}_3$ agent in this problem as well.

\item \textbf{Level3.} (Figure \ref{fig:resultG2}-(c)).
The $\mathscr{C}_3$ agent achieves the top performance in this most difficult setting.
It should be noted that, despite our original intent, the Stage-3 guidance was not beneficial in learning to solve the task, as indicated by the performance of Only 3 agent ($\mathscr{C}_4$).
Nevertheless, $\mathscr{C}_3$ agent showed the greatest robustness to such detrimental reward.
\end{itemize}

\subsubsection{Overall analysis on RWARE environments.}
\begin{figure}[t!]
\centering
    \includegraphics[width=0.9\columnwidth]{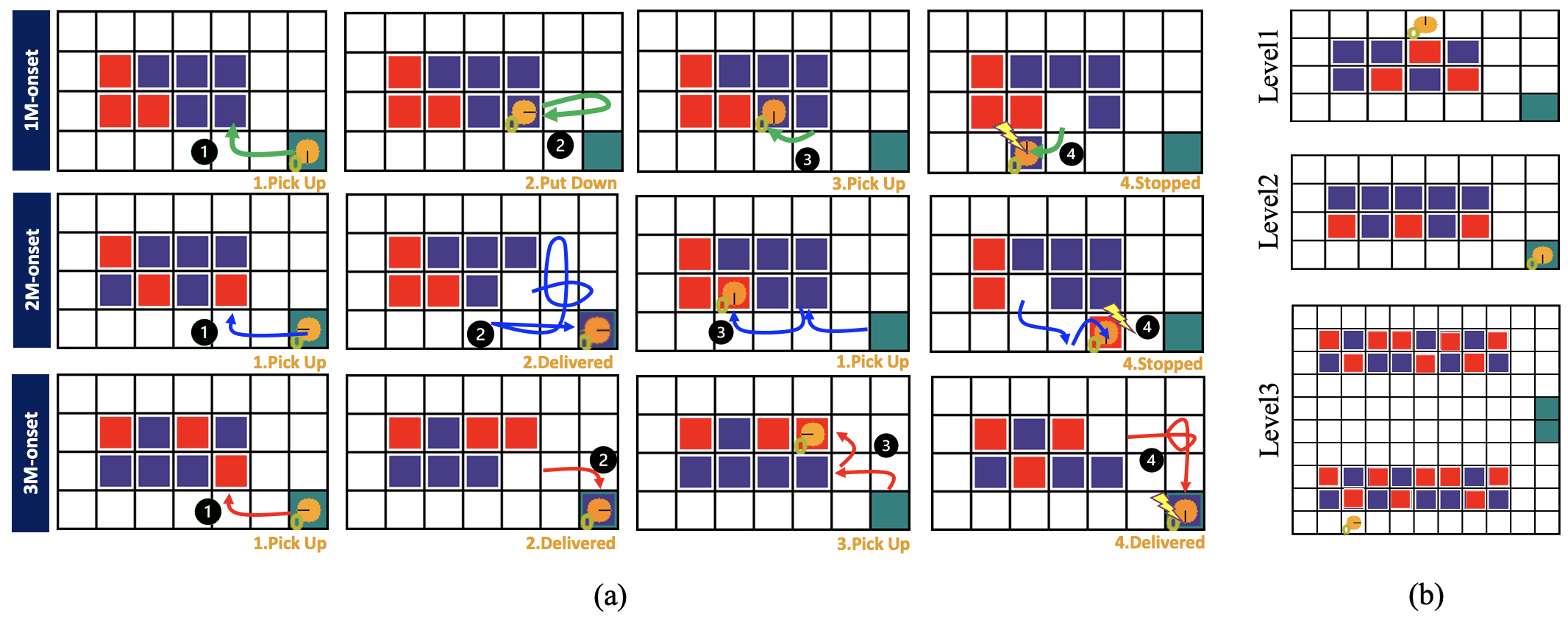}
    %\vskip -0.1in
    \caption{(a) Visualization of the trajectory for each agent policy in RWARE-Level1. (b) Three levels of RWARE environments.}
   \label{Rwaretrajectory}
   %\vskip -0.2in
\end{figure}

In Figure~\ref{Rwaretrajectory}, we illustrate the policy trajectories of various agents in RWARE-Level1. The goal, requested shelves, and non-requested shelves are marked as green, red, and blue squares respectively.
Here, the starting point of the agent is the same as the goal position (green).
The $\mathscr{C}_1$ agent picks incorrect shelves twice, and then halts exploration. The $\mathscr{C}_2$ agent initially picks the correct shelf and delivers it, but inefficiently explores the destination area in subsequent attempts. Unlike these two agents, the $\mathscr{C}_3$ agent delivers more shelves efficiently and experiences fewer subgoal failures. This highlights the impact of reward transition times on task performance. %and the propensity to fall into sub-optimal solutions.

\subsubsection{Importance of early free exploration.} We also observe that the $\mathscr{C}_1$ and $\mathscr{C}_2$ agents perform poorly when the first stage transition is given early, before the point of 3M training steps, as seen in Figure~\ref{fig:resultG2}. Therefore, we believe that a proper amount of exploration stage (Stage-1) given sparse reward is essential to successful learning. For Only 3 agent ($\mathscr{C}_5$), even with the richest reward, the agent fails completely in Level3. However, we found that our $\mathscr{C}_3$ agent is even robust to the unbeneficial reward through first stage transition in Toddler-inspired S2D setting.

\subsubsection{Limitations.} In the given context, the primary objective of this study was to investigate the influence of reward transition timings on performance metrics in tasks involving subgoals. The research documented the nuanced effects that the timing of these transitions exerts on overall task performance. However, it must be acknowledged that the establishment of an unequivocally optimal reward shaping strategy for tasks embedded with such subgoals remains elusive. Consequently, this delineates a clear avenue for future research, necessitating a comprehensive exploration of optimized reward shaping techniques. This includes, but is not limited to, a detailed examination of potential-based methods within this specialized environment. Such investigative efforts are projected to significantly improve the efficacy of toddler-inspired reward transition guidance mechanisms, ultimately leading to enhanced performance outcomes.

\subsection{C.2 Gridworld: Add-on Algorithms Experiments on 3D Loss Landscape}
\label{sec:gridworld}
\begin{figure}[!t]
\centering
    \includegraphics[width=0.9\textwidth]{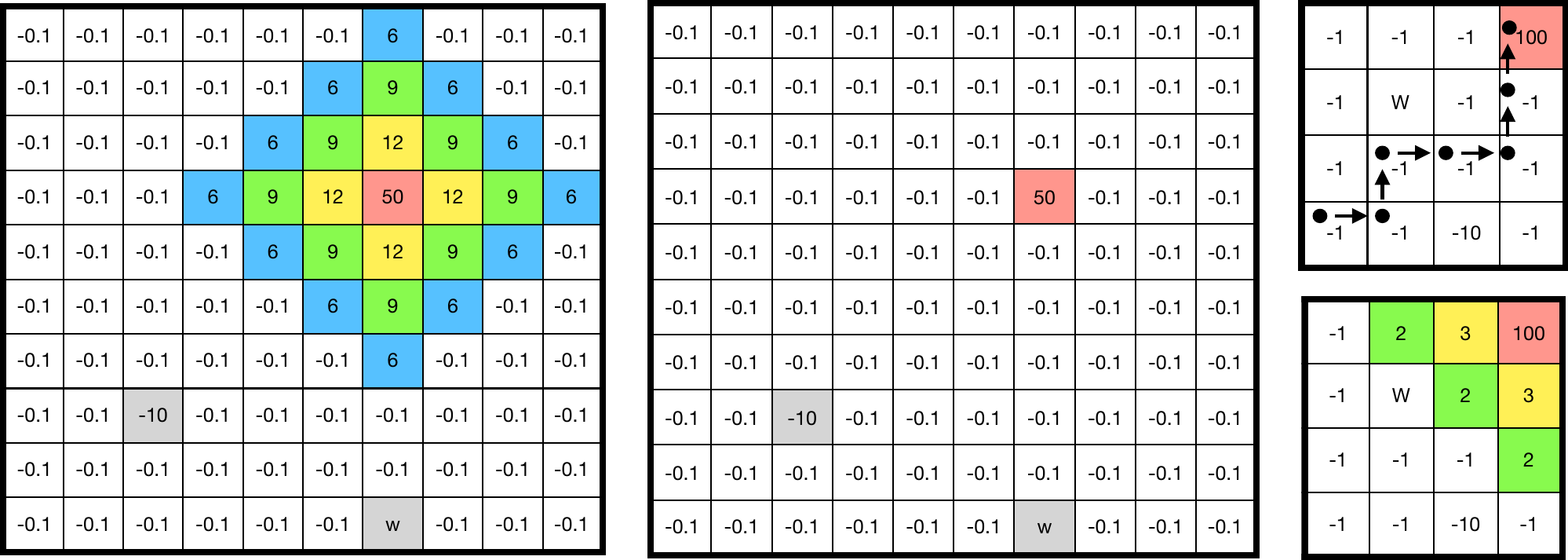}
    %\vskip -0.1in
    \caption{Gridworld-navigation task. \textbf{Left}: Potential-based dense reward environment of $10\times 10$ with \textit{PPO}. \textbf{Center}: sparse reward environment of $10\times 10$ with \textit{PPO}. \textbf{Right-Top}: sparse reward environment of $4\times 4$ with \textit{DQN}. \textbf{Right-Bottom}: Potential-based dense reward environment of $4\times 4$ with \textit{DQN}.}
   \label{fig:dense_sparse}
   % \vskip -0.25in
\end{figure}

We conducted 3D visualization of the policy landscape and experiments on performance in the Gridworld environment, which we selected due to its relative simplicity and minimal environmental factors. This setup provided a more straightforward context in which to observe the results according to both the S2D reward transition and the baseline, thereby offering clearer insights into the implications of our work. In the main text, we focused solely on visualizing the policy loss landscape using the SAC algorithm \cite{haarnoja2018soft}. To verify whether the smoothing effects still persist in other various algorithms, we conduced additional experiments using  the DQN~\cite{dqn2} and PPO~\cite{Schulman2017ProximalPO}.

\subsubsection{Environment setup.} The experimental environment is a gridworld, as shown in Figure~\ref{fig:dense_sparse}. The agent can perform actions of moving in four directions: up, down, left, and right. It receives a living penalty of $-0.1$ to encourage exploration. The experiment is conducted under two environment settings: (1) a fixed goal $4\times 4$ environment with DQN~\cite{dqn2} and (2) a random goal $10\times 10$ environment with PPO~\cite{Schulman2017ProximalPO}. Initially, we found an appropriate stage transition: $T=200$ for a fixed goal within a total of 1000 steps, and $T=5000$ for a random goal within a total of 100,000 steps. The neural network used for learning~\cite{dqn2, Schulman2017ProximalPO} consists of three fully connected layers with ReLU activation. The batch size is set to 128. In the case of PPO, we updated every 2 episodes. We employed the Cross-Density Visualizer strategy. 

%\subsubsection{Performance results.} In the gridworld experiment, the performance results are as follows. For DQN, the success rate of S2D is $72.23 \pm 2.29$, while Only Dense achieves $65.97 \pm 3.93$, and Only Sparse scores $61.00 \pm 2.06$. In the case of PPO, S2D shows a Success rate of $23.23 \pm 1.12$, Only Dense shows at $20.03 \pm 0.93$, Only Sparse reaches $18.9 \pm 0.96$, and D2S achieves $16.3 \pm 1.72$. We  also verify that the S2D reward transition outperforms all other baselines based on PPO and DQN algorithm.

%a notable smoothing effect, as depicted in Figures \ref{fig:girdppod2s}, \ref{fig:girdppos2d}, and \ref{fig:dense_sparse}.

%\subsubsection{3D Loss Landscape Analysis Post-Transition}
%\label{sec:strategy}

\subsubsection{The result of loss landscape in DQN and PPO algorithms.} 
\label{sec:strategy}
\begin{figure}[!hb]
\centering
    \includegraphics[width=\textwidth, height= 0.8\textheight, keepaspectratio]{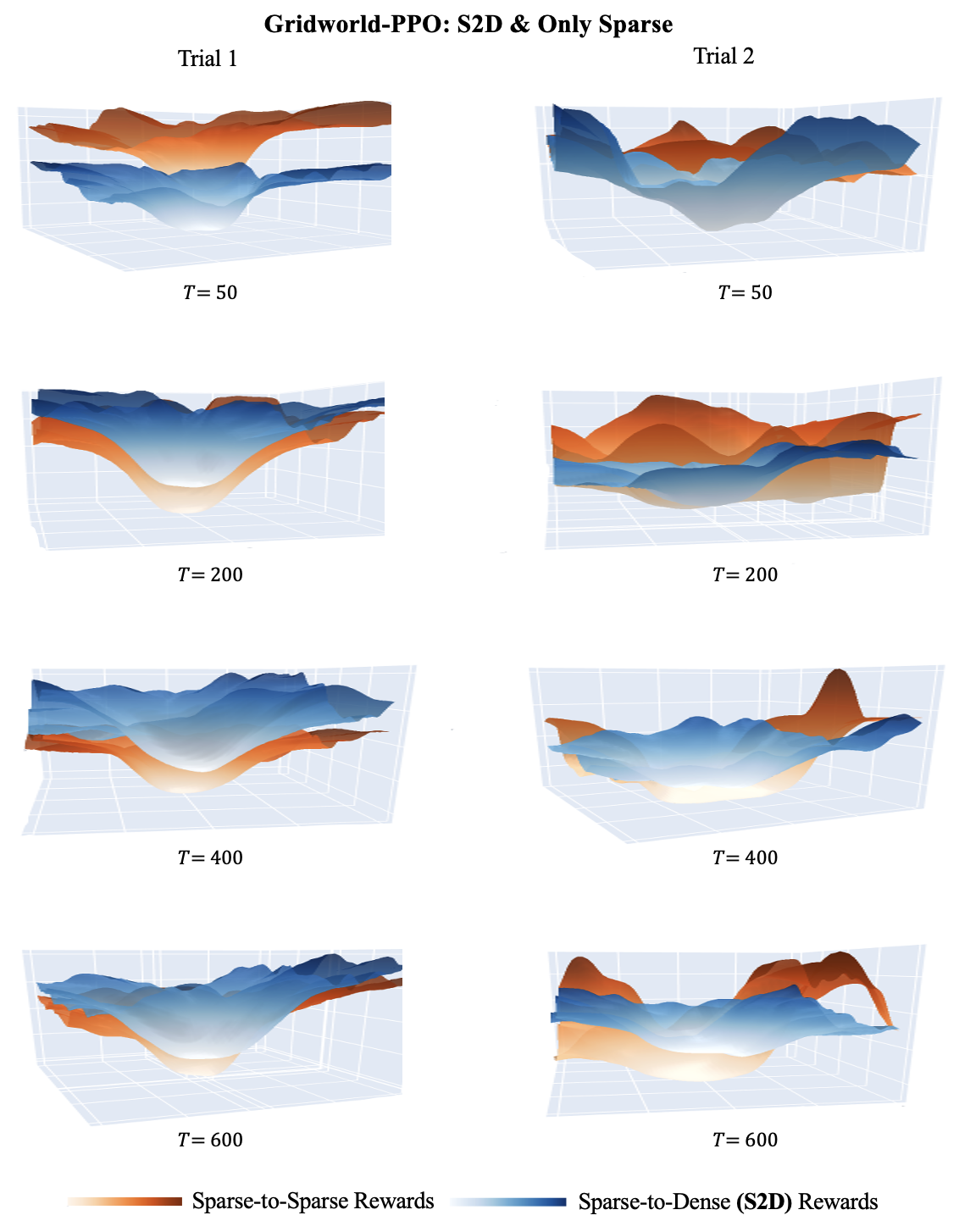}
    %\vskip -0.1in
    \caption{Visualization of the 3D policy loss landscape after stage-transitions in sparse-to-dense (\textcolor{blue}{Toddler-inspired S2D}, $\mathscr{C}_3$, blue) and sparse-to-sparse (Only Sparse, red). Here, we also observed notable smoothing effects (reduction in the depth of local minima) under Toddler-inspired S2D reward transition in PPO algorithm.}
   \label{fig:girdppos2d}
   % \vskip -0.25in
\end{figure}

\begin{figure}[!hb]
\centering
    \includegraphics[width=\textwidth, height= 0.8\textheight, keepaspectratio]{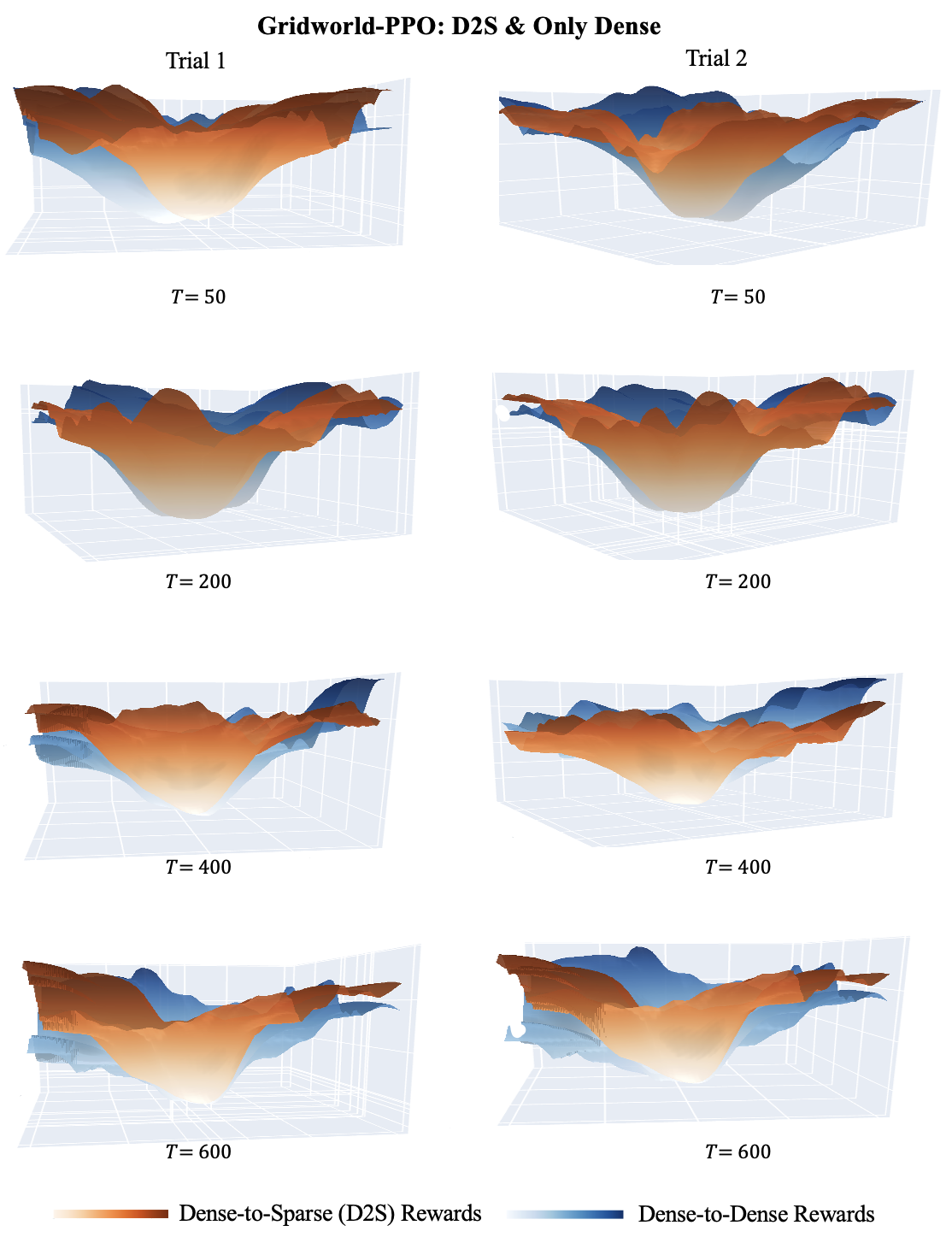}
    %\vskip -0.1in
    \caption{Visualization of the 3D policy loss landscape after stage-transitions in dense-to-sparse (D2S, $\mathscr{C}_3$, red) and dense-to-dense (Only Dense, blue). Similar to other D2S and Only Dense results in Appendix Section B, we also cannot observe notable smoothing effects consistent with the number of updates within Gridworld-PPO.}
   \label{fig:girdppod2s}
   % \vskip -0.25in
\end{figure}

\begin{figure}[!hb]
\centering
    \includegraphics[width=1\textwidth]{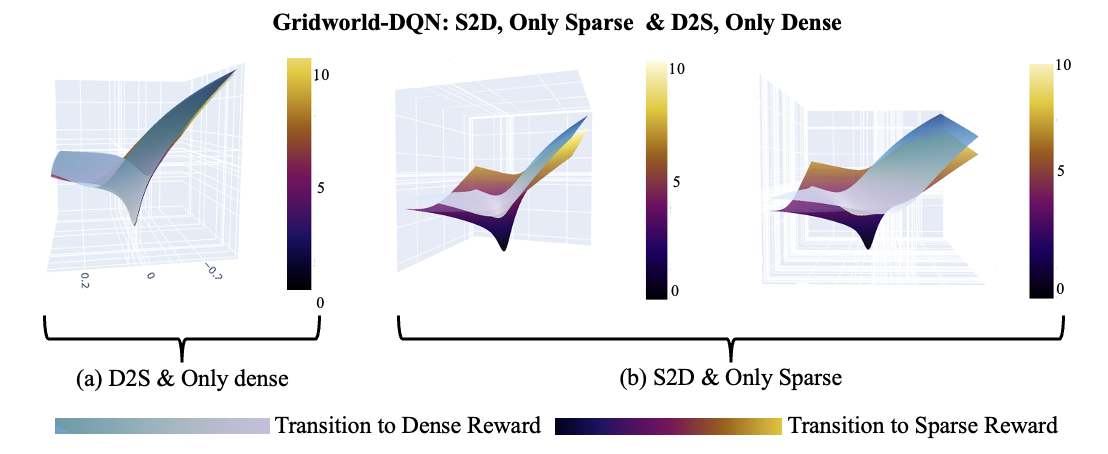}
    %\vskip -0.1in
    \caption{Visualization of the 3D Q-value loss landscape after reward transitions in Toddler-inspired S2D and other baselines. A smoothing effect is not apparent for the baselines (D2S, Only Dense, Only Sparse). However, under \textcolor{blue}{Toddler-inspired S2D} reward transition in (b) exclusively, a clear smoothing effect can be observed after transitioning to potential-based dense rewards by reducing the depth of local minima.}
   \label{fig:dense_sparse2}
   % \vskip -0.25in
\end{figure}

We visualized PPO for the policy loss landscape and DQN for the Q-function loss landscape. This is because, while DQN (Deep Q-Network) primarily aims to learn a Q-value function, observing the Q-value loss landscape can offer valuable insights into the network's learning dynamics and its adaptability to varied reward schemes. For both Gridworld-DQN and PPO, only the Toddler-inspired S2D reward transition exhibits a notable smoothing effect compared to other baselines, as depicted in Figures \ref{fig:girdppos2d}, \ref{fig:girdppod2s}, and \ref{fig:dense_sparse2}(b), achieving the highest performance. However, as seen in Figures \ref{fig:girdppod2s} and \ref{fig:dense_sparse2}-(a), we observe little to no difference between the dense and sparse loss landscapes as the number of updates increases. Therefore, we conclude that proceeding in a sparse-to-dense manner, i.e., using Toddler-inspired S2D reward transition, is an effective method for reducing the depth of local minima. It proves to be significantly more effective than the dense-to-sparse (D2S) or Only Dense approach.

\clearpage
\subsection{C.4 Detailed 3D Loss Landscape Visualizations for Gridworld-PPO and DQN Algorithm}
\subsection{Gridworld-PPO: Sparse-to-Dense (\textcolor{blue}{S2D}) \& Sparse-to-Sparse (Only Sparse)}
\clearpage
\subsection{Gridworld-PPO: Dense-to-Sparse (D2S) \& Dense-to-Dense (Only Dense)}

\clearpage
\subsection{Gridworld-DQN: \textcolor{blue}{S2D}, Only Sparse \& D2S, Only Dense}

\section*{Additional Future Works}

The approach examined in our research is inspired by the modeling of bio-inspired systems, particularly those based on toddler behaviors. Within this framework, our future research endeavors will focus on a deeper exploration of bio-plausible networks or models. We propose the following areas for future investigation, which we believe are pivotal for advancing our understanding at two critical junctures.

\begin{enumerate}

    \item \textbf{Reward Transition Patterns in SNN Based RL}: One promising avenue is the utilization of spiking neural networks (SNNs), which offer a more biologically realistic simulation of neural activity. Building upon bio-plausible SNN based RL research~\cite{qin2022low, zhang2022multi, zhang2022tuning}, we intend to investigate how SNNs can model and potentially enhance our understanding of reward transition dynamics in RL scenarios. By integrating toddler-inspired approaches into SNN research, we could provide novel insights into how rewards are processed and learned over time in SNN.
    \item \textbf{Exploring the Critical Period: Optimal Reward Transition Timing}: Another intriguing aspect is the initial period of free exploration in RL, which is crucial for learning effective strategies in our toddler-inspired reward transition. Through empirical experimentation, we were able to observe that the timing of the reward transition with proper guidance significantly affects subsequent performance and is a crucial parameter. This phenomenon can also be considered similar to the concept of the 'critical period' observed in infants. In human infants, appropriate training or proper guidance in the early stages of learning significantly impacts future learning. Moreover, the importance of suitable early education is evident from research on RL models~\cite{park2021toddler, de2022critical} and neural networks~\cite{achille2018critical} regarding this critical period. To advance this research area, our analysis, which includes the smoothing effect and low sharpness metrics of the S2D model, suggests that exploring the optimal timing for reward transition during the initial learning phase presents a promising avenue for future studies.
\end{enumerate}

We hope that our research will contribute to a broader understanding of neural network architectures and their learning mechanisms, bridging the gap between artificial intelligence and biological or cognitive modeling systems.

\newpage
\section{Pytorch-like Code for Loss Landscape Rendering}
\label{app:code}

\begin{algorithm}[h]
    \caption{Perpendicular Random Directions Generation}
    \begin{algorithmic}[1]
        \State $\text{rand\_direction\_list1} \gets []$
        \State $\text{rand\_direction\_list2} \gets []$
        \For{each $param$ in $network.parameters()$}
            \State $r \gets \text{torch.randn\_like}(param)$
            \State $r2 \gets \text{torch.randn\_like}(param)$
            \State $r \gets r - \left(\frac{\text{torch.dot}(r.\text{reshape}(-1), r2.\text{reshape}(-1))}{\| r2.\text{reshape}(-1) \|^2}\right) * r2$
            \State $r \gets \left(\frac{r}{\| r \|}\right) * \| param.\text{detach}() \|$
            \State $r2 \gets \left(\frac{r2}{\| r2 \|}\right) * \| param.\text{detach}() \|$
            \State $\text{rand\_direction\_list1}.\text{append}(\text{torch.nan\_to\_num}(r))$
            \State $\text{rand\_direction\_list2}.\text{append}(\text{torch.nan\_to\_num}(r2))$
        \EndFor
    \end{algorithmic}
\end{algorithm}

\begin{algorithm}[h]
    \caption{Deterministic Sampling}
    \begin{algorithmic}[1]
        \State \textbf{class} ReplayBuffer:
        \State \hspace{\algorithmicindent} \textbf{def} \_\_init\_\_(self, memory\_size):
        \State \hspace{\algorithmicindent} \hspace{\algorithmicindent} self.memory = deque(maxlen=memory\_size)
        \State \hspace{\algorithmicindent} \hspace{\algorithmicindent} self.experience = namedtuple("Experience",
        \State \hspace{\algorithmicindent} \hspace{\algorithmicindent}\hspace{\algorithmicindent} field\_names=["state", "action", "reward", "next\_state", "done", "episode"])
        \State \hspace{\algorithmicindent} \algcomment{ The rest is the same as a conventional replaybuffer.}
        \State \hspace{\algorithmicindent} \textbf{def} deterministic\_sample(self, size):
        \State \hspace{\algorithmicindent} \hspace{\algorithmicindent} idx = min(size-1, len(self.memory)-1)
        \State \hspace{\algorithmicindent} \hspace{\algorithmicindent} experiences = [self.memory[i] for i in range(len(self.memory)-1-idx, len(self.memory)-1)]
        \State
        \State \hspace{\algorithmicindent} \hspace{\algorithmicindent} states = np.vstack([e.state for e in experiences if e is not None])
        \State \hspace{\algorithmicindent} \hspace{\algorithmicindent} actions = np.vstack([e.action for e in experiences if e is not None])
        \State \hspace{\algorithmicindent} \hspace{\algorithmicindent} rewards = np.vstack([e.reward for e in experiences if e is not None])
        \State \hspace{\algorithmicindent} \hspace{\algorithmicindent} next\_states = np.vstack([e.next\_state for e in experiences if e is not None])
        \State \hspace{\algorithmicindent} \hspace{\algorithmicindent} dones = np.vstack([e.done for e in experiences if e is not None]).astype(np.bool\_)
        \State
        \State \hspace{\algorithmicindent} \hspace{\algorithmicindent} min\_epi = self.memory[0].episode
        \State \hspace{\algorithmicindent} \hspace{\algorithmicindent} max\_epi = self.memory[idx - 1].episode
        \State
        \State \hspace{\algorithmicindent} \hspace{\algorithmicindent} \textbf{return} states, actions, 
        \State \hspace{\algorithmicindent} \hspace{\algorithmicindent}\hspace{\algorithmicindent} np.squeeze(rewards, axis=1), next\_states, 
        \State \hspace{\algorithmicindent} \hspace{\algorithmicindent}\hspace{\algorithmicindent} np.squeeze(dones, axis=1)
    \end{algorithmic}
\end{algorithm}

\begin{algorithm}[h]
    \caption{Network Base}
    \begin{algorithmic}[1]
        \State \textbf{class} NetworkBase:
        \State \hspace{\algorithmicindent} \textbf{def} \_\_init\_\_(self):
        \State \hspace{\algorithmicindent} \hspace{\algorithmicindent} self.value\_net = ValueNetwork(state\_dim, hidden\_dim)
        \State \hspace{\algorithmicindent} \hspace{\algorithmicindent} self.target\_value\_net = ValueNetwork(state\_dim, hidden\_dim)
        \State
        \State \hspace{\algorithmicindent} \hspace{\algorithmicindent} self.soft\_q\_net1 = SoftQNetwork(state\_dim, action\_dim, hidden\_dim)
        \State \hspace{\algorithmicindent} \hspace{\algorithmicindent} self.soft\_q\_net2 = SoftQNetwork(state\_dim, action\_dim, hidden\_dim)
        \State \hspace{\algorithmicindent} \hspace{\algorithmicindent} self.policy\_net = PolicyNetwork(state\_dim, action\_dim, hidden\_dim)
    \end{algorithmic}
\end{algorithm}

\begin{algorithm}[h]
    \caption{Network Copy}
    \begin{algorithmic}[1]
        \Function{replica}{self}
            \State $result \gets \text{NetworkBase}(\text{self.device})$
            \For{each $target\_param, param$ in $\text{zip}(result.value\_net.parameters(), \text{self.value\_net.parameters()}):$}
                \State $target\_param.data.copy\_(param.data)$
                \Comment{Repeated for all networks.
                    }
            \EndFor
            \State \textbf{return} $result$
        \EndFunction
    \end{algorithmic}
\end{algorithm}

\begin{algorithm}[h]
    \caption{Loss Landscape Calculation for Policy}
    \begin{algorithmic}[1]
        \State $state, action, reward, next\_state, done \gets \text{replay\_buffer.deterministic\_sample()}$
        \State $xs \gets \text{torch.linspace}(-10, 10, \text{steps}=50)$
        \State $ys \gets \text{torch.linspace}(-10, 10, \text{steps}=50)$
        \State $xs, ys \gets \text{torch.meshgrid}(xs, ys, \text{indexing}='xy')$
        \State $zs \gets \text{torch.zeros\_like}(xs)$
        
        \For{$i$ in $\text{range}(xs.\text{shape}[0])$}
            \For{$j$ in $\text{range}(ys.\text{shape}[1])$}
                \State $new\_model \gets \text{self.replica()}$
                
                \For{$param, new\_param, rand\_dir1, rand\_dir2$ in \\ 
                \hspace{\algorithmicindent} \hspace{\algorithmicindent}
                \hspace{\algorithmicindent}\hspace{\algorithmicindent} zip$($self.policy\_net.parameters(), new\_model.policy\_net.parameters(), \\
                \hspace{\algorithmicindent} \hspace{\algorithmicindent} 
                \hspace{\algorithmicindent}
                \hspace{\algorithmicindent} self.rand\_direction\_list1, self.rand\_direction\_list2$)$ }

                    \State $transformed \gets param + xs[i][j] * rand\_dir1 + ys[i][j] * rand\_dir2$
                    \State $new\_param.\text{copy\_}(transformed)$
                \EndFor
                
                \State $new\_a, \text{log\_prob}, \epsilon, \text{mean}, \text{log\_std} \gets \text{new\_model.policy\_net.evaluate}(state)$
                \State $predicted\_new\_q\_value \gets \text{torch.min}($
                \State \hspace{\algorithmicindent} $\text{new\_model.soft\_q\_net1}(state, new\_a),$ 
                \State \hspace{\algorithmicindent} $\text{new\_model.soft\_q\_net2}(state, new\_a))$
                \State $zs[i][j] \gets (\text{log\_prob} - \text{predicted\_new\_q\_value}).\text{mean}().\text{item}()$
            \EndFor
        \EndFor
        
        \State \textbf{return} $xs.\text{numpy}(), ys.\text{numpy}(), zs.\text{numpy}()$
    \end{algorithmic}
\end{algorithm}

\begin{algorithm}[h]
    \caption{Cross-Density Visualizer}
    \begin{algorithmic}[1]
        \State $dense\_agent \gets \text{NetworkBase}(\text{device}=device)$
        \State $dense\_replay\_buffer \gets \text{ReplayBuffer}(\text{replay\_buffer\_size})$
        \For{$episode$ in $\text{trange}(transition\_episode):$}
            \State $state, \_ \gets \text{env.reset}(\text{seed}=seed)$
            \For{$step$ in $\text{range}(\text{max\_step}):$}
                \State $action \gets \text{env.action\_space.sample}()$
                \State $next\_state, reward, done, \_, \_$
                \State \hspace{\algorithmicindent} $= \text{env.step}(action)$
                \State $reward \gets \text{dense\_reward}(next\_state, reward)$
                \State $dense\_replay\_buffer.\text{push}(state, action, reward, next\_state, done, episode)$
                \State $state \gets next\_state$
                \State $dense\_agent.\text{update}(dense\_replay\_buffer)$
                \If{$done$}
                    \State \textbf{break}
                \EndIf
            \EndFor
            \State $\text{draw\_landscape}()$
        \EndFor
        \State
        \Comment{Cross-Density Running}
        \State $dense\_to\_sparse\_agent \gets dense\_agent.\text{replica}()$
        \State $dense\_to\_sparse\_replay\_buffer \gets \text{copy.deepcopy}(dense\_replay\_buffer)$
        \For{$episode$ in $\text{trange}(transition\_episode, \text{max\_episodes}):$}
            \State $state, \_ \gets \text{env.reset}(\text{seed}=seed)$
            \For{$step$ in $\text{range}(\text{max\_step}):$}
                \State $action \gets dense\_agent.policy\_net.\text{get\_action}(state).\text{detach}()$
                \State $next\_state, reward, done, \_, \_$
                \State \hspace{\algorithmicindent} $= \text{env.step}(action.\text{numpy}())$
                \State $reward \gets \text{dense\_reward}(next\_state, reward)$
                \State $dense\_replay\_buffer.\text{push}(state, action, reward, next\_state, done, episode)$
                \State $state \gets next\_state$
                \State $\_ \gets dense\_agent.\text{update}(dense\_replay\_buffer)$
                \If{$done$}
                    \State \textbf{break}
                \EndIf
            \EndFor
            \State
            \State $state, \_ \gets \text{env.reset}(\text{seed}=seed)$
            \For{$step$ in $\text{range}(\text{max\_step}):$}
                \State $action \gets dense\_to\_sparse\_agent.policy\_net.\text{get\_action}(state).\text{detach}()$
                \State $next\_state, reward, done, \_, \_$
                \State \hspace{\algorithmicindent} $= \text{env.step}(action.\text{numpy}())$
                \State $dense\_to\_sparse\_replay\_buffer.\text{push}(state, action, reward, next\_state, done, episode)$
                \State $state \gets next\_state$
                \State $\_ \gets dense\_to\_sparse\_agent.\text{update}(dense\_to\_sparse\_replay\_buffer)$
                \If{$done$}
                    \State \textbf{break}
                \EndIf
            \EndFor
            \State
            \State $\text{draw\_landscape}()$
        \EndFor
    \end{algorithmic}
\end{algorithm}

% \subfile{supplementary}
%\fi

\end{document}